%% file: main.tex
\documentclass[10pt,twocolumn,letterpaper]{article}
\usepackage{cvpr}
\input{preamble}
\definecolor{cvprblue}{rgb}{0.21,0.49,0.74}
\usepackage[pagebackref,breaklinks,colorlinks,allcolors=cvprblue]{hyperref}
\def\confName{CVPR}
\def\confYear{2027}
\title{Learning the Missing Derivative: Neural Multiple-Scattering Operators for Inverse Volume Rendering}
\author{Sipeng Chen \qquad Shibo Li\\
Florida State University}
\usepackage{xcolor}
\begin{document}
\twocolumn[{%
\renewcommand\twocolumn[1][]{#1}%
\maketitle
\input{sec/teaser}
}]

\input{sec/0_abstract}

\input{sec/1_intro}

\input{sec/2_related}

\input{sec/3_method}

\input{sec/4_experiments}
\input{sec/5_conclusion}
{\small\input{references_arxiv.tex}}
\input{sec/X_suppl}
\end{document}

%% file: preamble.tex
\usepackage{amsmath,amssymb,bm,booktabs,multirow,xcolor,graphicx,siunitx}
\usepackage[ruled,linesnumbered]{algorithm2e}
\SetCommentSty{algcomment}   
\usepackage{tikz}\usetikzlibrary{arrows.meta,positioning,calc}   
\usepackage{placeins}   
\usepackage{xspace}   
\definecolor{cOurs}{HTML}{C0392B}
\newcommand{\vx}{\mathbf{x}}                            
\newcommand{\vo}{\mathbf{o}}                            
\newcommand{\vd}{\mathbf{d}}                            
\newcommand{\vw}{\boldsymbol{\omega}}                   
\newcommand{\wl}{\boldsymbol{\omega}_{l}}               
\newcommand{\wo}{\boldsymbol{\omega}_{o}}               
\newcommand{\sigt}{\sigma_{t}}                          
\newcommand{\sigs}{\sigma_{s}}                          
\newcommand{\Tcam}{T_{\mathrm{cam}}}                    
\newcommand{\Tl}{T_{l}}                                 
\newcommand{\taul}{\tau_{l}}                            
\newcommand{\Lms}{L^{(\geq 1)}}                         
\newcommand{\Sop}{S_{\phi}}                             
\newcommand{\Smc}{S_{\mathrm{MC}}}                      
\newcommand{\Mcal}{\mathcal{M}}                         
\newcommand{\sg}{\operatorname{sg}}                     
\newcommand{\mJ}{\mathbf{J}}                            
\newcommand{\Jcam}{\mathbf{J}_{\mathrm{cam}}}           
\newcommand{\Jdir}{\mathbf{J}_{\mathrm{direct}}}        
\newcommand{\Jms}{\mathbf{J}_{\mathrm{multi}}}          
\newcommand{\Jmsphi}{\mathbf{J}_{\mathrm{multi},\phi}}  
\newcommand{\vI}{\mathbf{I}}                            
\newcommand{\vr}{\mathbf{r}}                            
\newcommand{\vy}{\mathbf{y}}                            
\newcommand{\vv}{\mathbf{v}}                            
\newcommand{\vdelta}{\boldsymbol{\delta}}               
\newcommand{\FS}{Frozen Source\xspace}                  
\newcommand{\DO}{Detached Operator\xspace}              
\input{tables/gpu_hours_macros}   
\input{tables/opfield_macros}     
\input{tables/dtime_macros}       
\usepackage{zref-user,zref-abspage,zref-savepos}          

%% file: tables/gpu_hours_macros.tex
\newcommand{\gpuhDev}{881}
\newcommand{\gpuhPoolOne}{283}
\newcommand{\gpuhPoolTwo}{150}
\newcommand{\gpuhPoolThree}{220}
\newcommand{\gpuhTotal}{1{,}534}

%% file: tables/opfield_macros.tex
\newcommand{\opfValRange}{2.2--4.9\%}

\newcommand{\opfCosRange}{0.95--0.99}

\newcommand{\opfFrozenRange}{0.1--18.1\%}
\newcommand{\opfImgCosRange}{0.91--0.99}

\newcommand{\opfFigVal}{2.9\%}

\newcommand{\opfFigCos}{0.99}

\newcommand{\opfFigPhotonsRef}{$9.8\times10^{9}$}
\newcommand{\opfFigPhotonsFd}{$7.4\times10^{9}$}

\newcommand{\opfFigFarShare}{31\%}
\newcommand{\opfFigFarCos}{0.99}
\newcommand{\opfSecantDiffMax}{0.002}

\newcommand{\opfFrozenMed}{1.2\%}
\newcommand{\opfPhRefRange}{1.0--2.3}
\newcommand{\opfPhFdRange}{1.5--3.4}
\newcommand{\opfRefNoiseMax}{1.0\%}
\newcommand{\opfLinMin}{0.9995}
\newcommand{\opfPrecMin}{0.99999}

%% file: tables/dtime_macros.tex
\newcommand{\dtGPU}{B200}
\newcommand{\dtNetBf}{13\,ms}
\newcommand{\dtNetFp}{33\,ms}
\newcommand{\dtInputs}{7\,ms}

\newcommand{\dtOurs}{77\,ms}
\newcommand{\dtOursRange}{75--79\,ms}
\newcommand{\dtOursCalls}{180}
\newcommand{\dtFwd}{38\,ms}
\newcommand{\dtFwdBf}{18\,ms}
\newcommand{\dtOursErr}{0.16}

\newcommand{\dtRefNoise}{0.09}
\newcommand{\dtRefPhotons}{$7.4\times10^{9}$}
\newcommand{\dtRef}{45\,min}

\newcommand{\dtRate}{$1.1\times10^{7}$}
\newcommand{\dtMinPhotons}{$10^{7}$}
\newcommand{\dtTrain}{8.5\,h}
\newcommand{\dtRateRange}{$1.1$--$2.6\times10^7$}
\newcommand{\dtLbPhotons}{$3\times10^{9}$}
\newcommand{\dtLb}{9.1\,min}
\newcommand{\dtLbRatio}{7{,}100}

\newcommand{\dtMatch}{23\,min}
\newcommand{\dtMatchRatio}{17{,}000}
\newcommand{\dtMatchRatioExact}{17{,}600}
\newcommand{\dtMatchErr}{0.162--0.163}
\newcommand{\dtOursErrThree}{0.163}

%% file: sec/teaser.tex
\begin{center}
  \centering
  \captionsetup{type=figure}
  \includegraphics[width=\textwidth]{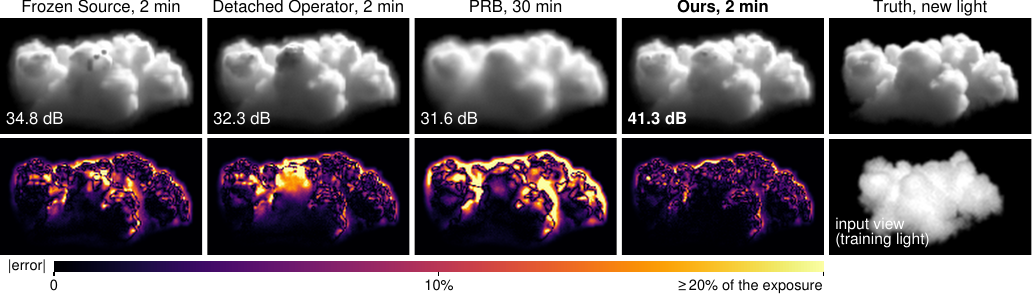}
  \captionof{figure}{\textbf{Relighting a cloud under novel illumination} -- A thick cloud reconstructed from 32 photographs under one
    light (bottom right: an input photograph) and relit under a new light. Differentiating through our learned
    multiple-scattering operator relights it most accurately after 2\,min, whereas the same pipeline without the operator
    (\FS~\cite{levis2015airborne}) and with the operator's values but without its derivative (\DO) misplace density in
    the interior, and differentiable path tracing (PRB~\cite{vicini2021prb}) is still blurred after 30\,min. Reported PSNR is evaluated over three novel lights and two unseen views. The lower row shows absolute error for the displayed relighting.}
  \label{fig:teaser}
\end{center}

%% file: sec/0_abstract.tex
\begin{abstract}
Reconstructing optically thick media from photographs requires modeling multiple scattering and its response to changes in extinction. Frozen-source tomography discards this derivative; differentiable path tracing estimates it through expensive, noisy sampling. We instead predict the multiple-scattering field with a neural operator and differentiate through it during inverse rendering, while computing transmittance and single scattering directly. A further challenge arises during optimization: an operator that predicts the field accurately on held-out media can still produce poor derivatives at the density states visited by the optimizer. We address this by fine-tuning on states generated during inversions driven by the operator, using Monte Carlo scattering-field supervision. On held-out optically thick clouds, our method improves relighting over frozen-source tomography at equal time and outperforms differentiable path tracing given substantially more time. Training on inversion trajectories also improves derivative alignment at visited states and reduces optimization drift.
\end{abstract}

%% file: sec/1_intro.tex
\section{Introduction}
\label{sec:intro}

\emph{``Light travels in a straight line.''}
This simple principle has shaped modern 3D reconstruction.
NeRF~\cite{mildenhall2020nerf} and 3D Gaussian Splatting~\cite{kerbl20233dgs} reconstruct scenes by
accumulating contributions along camera rays, typically
without explicitly modeling multiple scattering. But in optically thick media such as clouds and
smoke, light changes direction repeatedly before reaching the
camera. A pixel can therefore depend on light scattered far outside
its camera ray.

A reconstruction may fit the input views yet contain errors
in its interior extinction that only become visible under
new lighting. To reconstruct these media from photographs and
relight them under new illumination, we must account for
multiple scattering. For gradient-based inverse rendering,
we must also differentiate this transport with respect to
the medium's density.

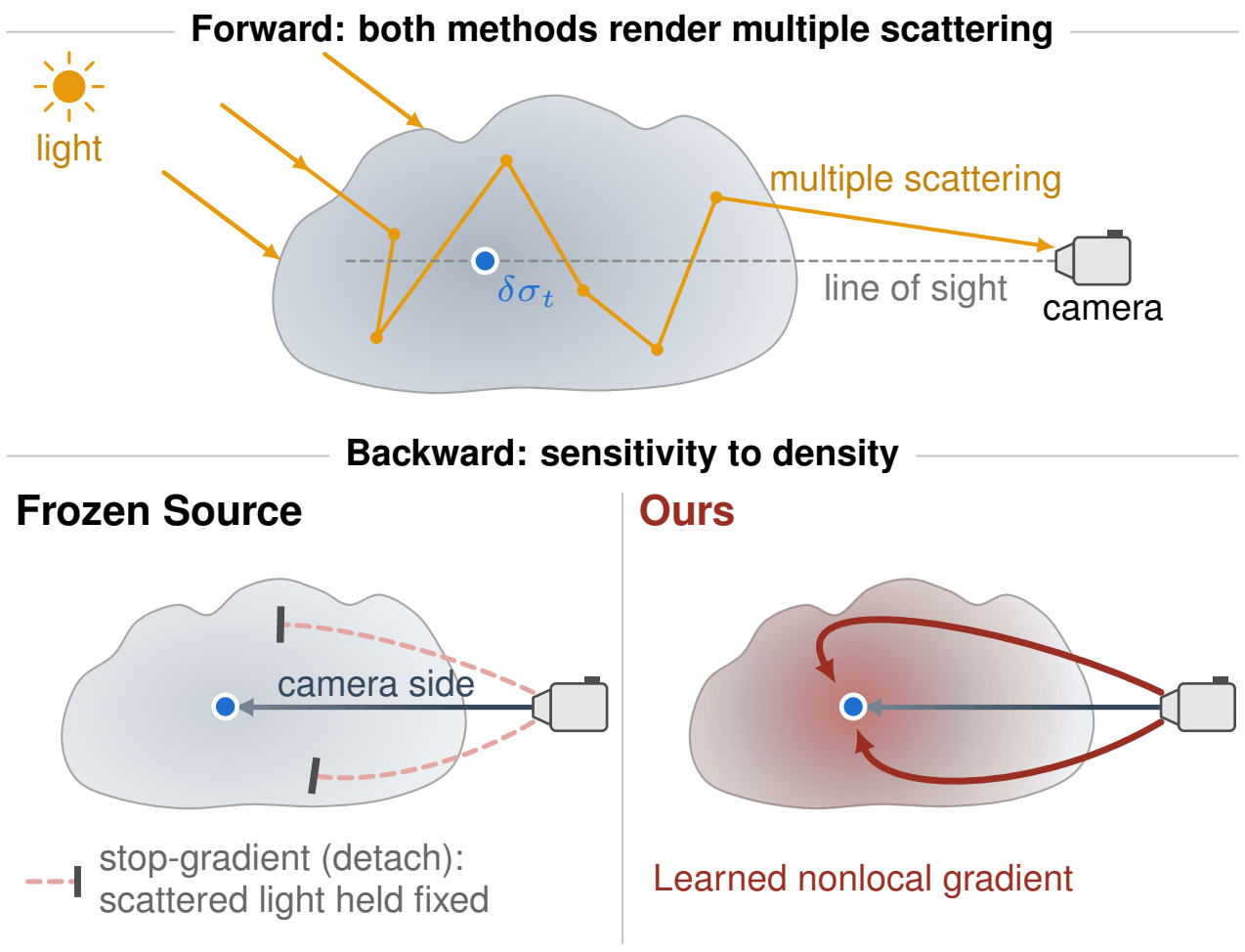
\begin{figure}[t]
  \centering
  \input{figures/concept_tikz}
  \caption{\textbf{Why the missing derivative matters} --
    In the forward pass both methods render all orders of
    scattering. In the backward pass,
    \FS~\cite{levis2015airborne} holds the scattered light
    fixed (stop-gradient): the gradient reaches a local density
    change $\delta\sigt$ only along its line of sight
    (camera side). Ours also differentiates through the neural
    operator $G_\phi$, whose learned nonlocal gradient (red)
    captures how $\delta\sigt$ redistributes the scattered
    light far beyond that line (the exact direct-beam term
    is not drawn).}
  \label{fig:concept}
\end{figure}

Computing this derivative is a central bottleneck. In optically
thick media, the multiple-scattering response to density
changes can be substantially larger than the camera-side
response. Differentiable path tracing estimates the full gradient through sampled light
paths~\cite{vicini2021prb,nimierdavid2022drt}, but the
computation is expensive and subject to Monte Carlo
noise~\cite{nimierdavid2022drt,leonard2024sps,nsampi2026nrt}.
Frozen-source tomography takes a cheaper
approach~\cite{levis2015airborne}: it periodically updates the
scattered-light field but treats it as constant when optimizing
density (\cref{fig:concept}). This approximation omits the
nonlocal multiple-scattering derivative, potentially producing
incorrect interior density even when the observed images are
well matched. Such errors become apparent when the reconstructed medium is rendered under new lighting
(\cref{fig:teaser}).
To eliminate them, can we obtain an accurate multiple-scattering derivative without repeatedly performing expensive Monte Carlo differentiation?

Our approach predicts the multiple-scattering source with a
neural operator and differentiates through that prediction
during inversion. The resulting derivative captures how the
scattered-light field responds to changes in density.
Transmittance and single scattering remain physically
computed; the neural operator supplies the expensive
light-side response (\cref{fig:overview}).

However, realizing this hybrid solver requires more than standard offline training: accurate field prediction does not guarantee accurate
derivatives, especially inside an inversion loop. Each approximate gradient
determines the next density state, which in turn becomes the
next input to the operator. As optimization proceeds, these
states may depart from the training distribution, allowing
derivative errors to accumulate and redirect subsequent
updates. The operator must therefore remain accurate along the
trajectory its own gradients induce.

To address this problem, we train the operator on the states
visited by inversions that use the operator itself. Starting
from a pretrained operator, we collect intermediate density
states generated during inversion, obtain Monte Carlo
supervision for their scattering fields, and fine-tune the
operator on these states. This improves derivative accuracy on visited states and
reduces optimization drift.

On 12 held-out optically thick clouds, our method improves relighting over frozen-source tomography on every medium, with a median gain of 6.8\,dB after two minutes. It also outperforms differentiable path tracing after thirty minutes (\cref{sec:exp:main}), while training on inversion-visited states substantially improves optimization stability (\cref{sec:exp:trajectory}).

Our contributions are:
\begin{itemize}
  \item \textbf{Learning the missing derivative.}
  We introduce a neural multiple-scattering operator that
  supplies the otherwise expensive light-side derivative
  through automatic differentiation, enabling efficient
  gradient-based inverse volume rendering.

  \item \textbf{Learning the operator for inversion.}
  We identify the limitations of conventional operator
  accuracy for inverse rendering and introduce training on
  inversion-visited states to improve derivative reliability
  and optimization stability.
\end{itemize}

%% file: figures/concept_tikz.tex
\begin{tikzpicture}[font=\sffamily\scriptsize, line cap=round, line join=round,
  lightray/.style={draw=cLight, line width=0.85pt, -{Latex[length=3.6pt,width=3.2pt]}},
  photon/.style={draw=cLight, line width=0.75pt},
  ms/.style={draw=cOurs!80!black, line width=1.3pt, -{Latex[length=4.6pt,width=4.2pt]}, shorten >=1.2pt},
  msoff/.style={draw=cOurs!45, line width=0.95pt, dash pattern=on 2.4pt off 1.7pt},
  ttl/.style={font=\sffamily\footnotesize\bfseries, inner sep=0pt},
  hdr/.style={font=\sffamily\scriptsize\bfseries, fill=white, inner xsep=3pt, inner ysep=1pt},
  note/.style={font=\sffamily\scriptsize, inner sep=1pt, align=center}]
\definecolor{cLight}{HTML}{E69A0C}%
\definecolor{cCam}{HTML}{34495E}%
\definecolor{cPert}{HTML}{1F6FD1}%
\definecolor{cCloudIn}{HTML}{AAB4C0}%
\definecolor{cCloudOut}{HTML}{E6EAEF}%
\def\cfShape{plot[smooth cycle, tension=0.62] coordinates {
  (-1.00,-0.10) (-0.92,0.16) (-0.74,0.30) (-0.62,0.47) (-0.42,0.55) (-0.26,0.50) (-0.10,0.63) (0.10,0.68)
  (0.30,0.62) (0.44,0.55) (0.60,0.60) (0.78,0.50) (0.90,0.33) (1.00,0.12) (1.02,-0.10) (0.92,-0.30)
  (0.64,-0.42) (0.30,-0.46) (-0.05,-0.45) (-0.40,-0.47) (-0.72,-0.42) (-0.93,-0.32)}}%
\newcommand{\cfCloud}[4][100]{
  \begin{scope}[shift={(#2,#3)}, scale=#4]
    \begin{scope}\clip \cfShape;
      \shade[inner color=cCloudIn!#1!white, outer color=cCloudOut!#1!white] (-0.18,0.04) ellipse (1.25 and 0.85);
    \end{scope}
    \draw[black!35, line width=0.4pt] \cfShape;
  \end{scope}}%
\newcommand{\cfCamera}[2]{
  \begin{scope}[shift={(#1,#2)}]
    \fill[black!10, draw=black!70, line width=0.5pt] (0,-0.065) -- (0.12,-0.12) -- (0.12,0.12) -- (0,0.065) -- cycle;
    \fill[black!10, draw=black!70, line width=0.5pt, rounded corners=1pt] (0.12,-0.16) rectangle (0.50,0.16);
    \fill[black!70] (0.35,0.16) rectangle (0.45,0.21);
  \end{scope}}%
\newcommand{\cfPert}[2]{\fill[white] (#1,#2) circle (0.095); \fill[cPert] (#1,#2) circle (0.064);}%
\newcommand{\cfSun}[2]{\fill[cLight] (#1,#2) circle (0.11);
  \foreach \a in {0,45,...,315} {\draw[cLight, line width=0.6pt] ($(#1,#2)+(\a:0.16)$) -- ($(#1,#2)+(\a:0.23)$);}}%
\useasboundingbox (0,0) rectangle (8.33,6.30);
\draw[black!22, line width=0.4pt] (0,6.17) -- (8.33,6.17);
\node[hdr] at (4.165,6.17) {Forward: both methods render multiple scattering};
\cfCloud{3.55}{4.55}{1.75}
\cfSun{0.42}{5.80}
\node[note, text=cLight!85!black, anchor=north] at (0.42,5.52) {light};
\draw[lightray] (2.15,6.02) -- (2.83,5.51);
\draw[lightray] (1.06,5.22) -- (1.86,4.62);
\draw[photon] (1.455,5.675) -- (2.62,4.80) -- (2.50,4.10) -- (3.38,5.30) -- (3.90,4.42) -- (4.40,4.02) -- (4.80,5.05);
\draw[photon, -{Latex[length=3.6pt,width=3.2pt]}] (1.455,5.675) -- (2.06,5.22);
\draw[photon, -{Latex[length=3.6pt,width=3.2pt]}] (4.80,5.05) -- (7.10,4.71);
\foreach \x/\y in {2.62/4.80, 2.50/4.10, 3.38/5.30, 3.90/4.42, 4.40/4.02, 4.80/5.05} {
  \fill[cLight] (\x,\y) circle (0.042);}
\draw[black!45, line width=0.5pt, dash pattern=on 1.5pt off 1.2pt] (2.30,4.62) -- (7.10,4.62);
\cfPert{3.235}{4.62}
\node[note, text=cPert, anchor=north west] at (3.27,4.56) {$\delta\sigt$};
\cfCamera{7.10}{4.62}
\node[note, anchor=north] at (7.42,4.40) {camera};
\node[note, text=cLight!85!black, anchor=south] at (6.15,5.00) {multiple scattering};
\node[note, text=black!55, anchor=north] at (6.15,4.58) {line of sight};
\draw[black!22, line width=0.4pt] (0,3.30) -- (8.33,3.30);
\node[hdr] at (4.165,3.30) {Backward: sensitivity to density};
\draw[black!22, line width=0.4pt] (4.165,0) -- (4.165,3.05);
\node[ttl, anchor=north west] at (0.05,3.05) {Frozen Source};
\node[ttl, anchor=north west, text=cOurs!80!black] at (4.27,3.05) {Ours};
\foreach \ox in {0, 4.25} {\cfCloud[62]{\ox+1.72}{1.55}{1.35}}
\begin{scope}
  \clip[shift={(5.97,1.55)}, scale=1.35] \cfShape;
  \foreach \i in {1,...,60} {\fill[cOurs, fill opacity=0.0122] (5.727,1.604) circle ({1.5*pow(\i/60,1.6)});}
\end{scope}
\foreach \ox in {0, 4.25} {
  \draw[cCam, line width=1.0pt] (\ox+3.56,1.60) -- (\ox+3.06,1.60);
  \foreach \k in {0,...,10} {
    \pgfmathsetmacro{\xa}{\ox+3.06-0.13*\k}
    \pgfmathsetmacro{\cc}{100-3*\k}
    \draw[cCam!\cc!white, line width=1.0pt] (\xa,1.60) -- (\xa-0.13,1.60);}
  \draw[cCam!68!white, line width=1.0pt, -{Latex[length=4.2pt,width=3.8pt]}, shorten >=1.2pt]
    (\ox+1.63,1.60) -- (\ox+1.477,1.604);
  \cfCamera{\ox+3.56}{1.60}}
\draw[msoff, -{Bar[width=7pt, line width=1.4pt, color=black!70]}]
  (3.560,1.700) .. controls (3.043,1.948) and (2.311,2.150) .. (1.826,2.160);
\draw[msoff, -{Bar[width=7pt, line width=1.4pt, color=black!70]}]
  (3.560,1.500) .. controls (3.056,1.200) and (2.476,1.080) .. (2.053,1.140);
\draw[ms] (7.81,1.70) .. controls (6.87,2.15) and (5.22,2.45) .. (5.63,1.72);
\draw[ms] (7.81,1.50) .. controls (6.97,1.00) and (5.92,1.00) .. (5.735,1.50);
\foreach \ox in {0, 4.25} {\cfPert{\ox+1.477}{1.604}}
\node[note, text=cCam, anchor=south] at (2.50,1.63) {camera side};
\draw[msoff, -{Bar[width=6pt, line width=1.3pt, color=black!70]}] (0.14,0.42) -- (0.50,0.42);
\node[note, text=black!60, anchor=west, align=left] at (0.58,0.42) {stop-gradient (detach):\\scattered light held fixed};
\node[note, text=cOurs!80!black, anchor=west, align=left] at (4.33,0.42) {Learned nonlocal gradient};
\end{tikzpicture}%

%% file: sec/2_related.tex
\section{Related Work}
\label{sec:related}

\begin{figure*}[t]
  \centering
  \resizebox{\textwidth}{!}{\input{figures/overview_tikz}}
  \caption{\textbf{Method overview} --
    (a)~The inversion renders the training views with exact
    transmittance and single scattering, and with the
    multiple-scattering source predicted by $G_\phi$.
    Differentiating through the operator supplies
    $\Jmsphi$ for the density update; an occasional Monte
    Carlo correction $\Delta$ adjusts the values only.
    (b)~The operator is pretrained on generic media,
    fine-tuned for the target regime, and further trained
    on density states visited during inversion.}
  \label{fig:overview}
\end{figure*}
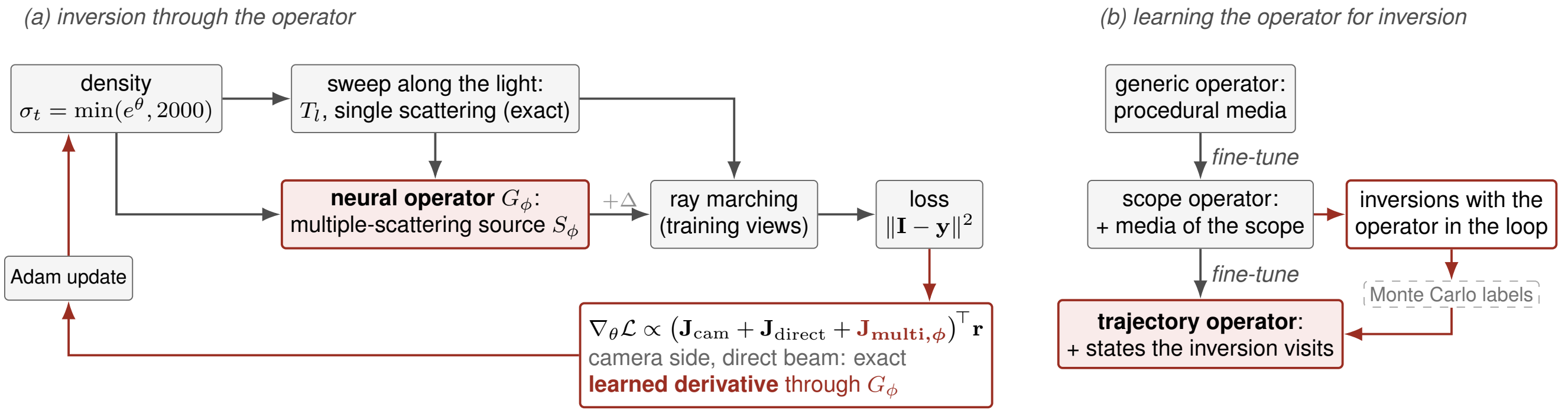

\paragraph{Differentiable volume rendering and cloud tomography.}
Inverse rendering of participating media requires differentiating
volumetric light transport. Monte Carlo methods, including
adjoint transport and path replay backpropagation (PRB), estimate
the full gradient by tracing light
paths~\cite{gkioulekas2013inverse,vicini2021prb}.
Variance reduction and path reuse reduce the
cost~\cite{nimierdavid2022drt,czerninski2023pars}, but
reconstructing dense scattering volumes remains expensive.

Early cloud tomography methods reconstruct cloud extinction using an SHDOM forward model while holding the in-scattered source fixed during differentiation~\cite{levis2015airborne}.
This frozen-source approximation has since been used with
both deterministic and Monte Carlo transport
models~\cite{levis2017microphysics,holodovsky2016insitu}.
Later methods also differentiate single scattering or full
light paths~\cite{levis2020pyshdom,loeub2020monotonicity}, and learned inverses map images to extinction
directly~\cite{sdechen2021deepct,ronen2022vipct,ronen2025probct}.
Yet freezing the scattered source removes its nonlocal
response to density changes, which dominates the camera-side
response in optically thick media. Our operator approximates this response and provides its derivative through automatic differentiation, avoiding repeated Monte Carlo gradient estimation.

\paragraph{Neural multiple scattering and neural operators.}
Neural networks have been used to predict in-scattered
radiance for cloud rendering, including models that
generalize to unseen
media~\cite{kallweit2017deepscattering,hu2023mrpnn}.
Other approaches, such as NRPM, photon field networks,
and radiance caches, fit or store light transport for
individual scenes~\cite{zheng2021nrpm,bauer2024photon,
muller2021nrc}.

Neural operators~\cite{li2021fourier} have been trained to predict fluence and
radiance fields, including from noisy Monte Carlo labels~\cite{cao2026ptno,carne2026fourier}.
A related approach in photoacoustic tomography uses
a pretrained neural operator to predict light fluence,
but holds that field fixed when updating
absorption~\cite{liang2023pat}.
Our method instead differentiates the predicted
multiple-scattering field with respect to extinction
and uses its Jacobian directly in inverse rendering.
A single pretrained operator predicts the scattering source for unseen media, and the resulting field can be differentiated and reused across camera views during inversion.

\paragraph{Learned surrogates for inverse optimization.}
A surrogate used inside an optimizer influences the
states at which it will be evaluated next.
Model-management methods correct surrogate values and
gradients during optimization~\cite{alexandrov1998trust},
while solver-in-the-loop and rollout-based training expose
learned models to states produced by their own
predictions~\cite{um2020solverintheloop,brandstetter2022mp}.
Adaptive and derivative-informed neural operators further
improve surrogate accuracy where it matters for
optimization~\cite{gao2023adaptive,oleary2022dino}.
Neural operators can also be trained directly on optimization trajectories using supervision on the objective gradient~\cite{cheng2025accelerating}.

We take an operator-in-the-loop approach to inverse
volume rendering: the operator supplies the
multiple-scattering derivative while the camera-side
and direct-light terms remain physically computed.
Following the idea of DAgger~\cite{ross2011dagger},
we collect the density states visited by inversions
driven by the operator and fine-tune it using Monte Carlo
scattering-field labels.

%% file: figures/overview_tikz.tex
\begin{tikzpicture}[
  font=\sffamily\footnotesize, >=Latex,
  box/.style={draw, rounded corners=1.5pt, align=center, inner sep=3pt, minimum height=8.5mm},
  op/.style={box, fill=cOurs!10, draw=cOurs!80!black, line width=0.8pt},
  mc/.style={draw=black!35, dashed, rounded corners=1.5pt, align=center, inner sep=2pt, text=black!55,
             font=\sffamily\scriptsize},
  ex/.style={box, fill=black!4, draw=black!60},
  grad/.style={->, thick, cOurs!80!black},
  fwd/.style={->, thick, black!70},
  lbl/.style={font=\sffamily\footnotesize\itshape, text=black!70}]
\node[lbl, anchor=west] at (-1.3, 1.0) {(a) inversion through the operator};
\node[ex] (theta) at (0, 0) {density\\$\sigma_t=\min(e^{\theta},2000)$};
\node[ex] (ss) at (4.0, 0) {sweep along the light:\\$T_l$, single scattering (exact)};
\node[op] (G) at (4.0, -1.45) {\textbf{neural operator} $G_\phi$:\\multiple-scattering source $S_\phi$};
\node[ex] (render) at (7.75, -1.45) {ray marching\\(training views)};
\node[ex] (loss) at (10.2, -1.45) {loss\\$\|\mathbf{I}-\mathbf{y}\|^2$};
\draw[fwd] (theta) -- (ss);
\draw[fwd] (theta.south) |- (G.west);
\draw[fwd] (ss.south) -- (G.north);
\draw[fwd] (ss.east) -| (render.north);
\draw[fwd] (G) -- node[above, inner sep=1.5pt, font=\sffamily\scriptsize, text=black!45] {$+\Delta$} (render);
\draw[fwd] (render) -- (loss);
\node[box, draw=cOurs!80!black, line width=0.8pt, align=left, anchor=north east] (J) at (11.0, -2.55) {%
  $\nabla_\theta\mathcal{L}\propto\big(\mathbf{J}_{\mathrm{cam}}+\mathbf{J}_{\mathrm{direct}}
   +\textcolor{cOurs!80!black}{\boldsymbol{\mathbf{J}_{\mathrm{multi},\phi}}}\big)^{\!\top}\mathbf{r}$\\[1pt]
  \textcolor{black!60}{camera side, direct beam: exact}\\
  \textcolor{cOurs!80!black}{\textbf{learned derivative} through $G_\phi$}};
\draw[grad] (loss.south) -- (loss.south |- J.north);
\node[ex, minimum height=5.5mm, inner sep=2pt, font=\sffamily\scriptsize] (adam) at (-0.6, -2.25) {Adam update};
\draw[grad] (J.west) -| (adam.south);
\draw[grad] (adam.north) -- (adam.north |- theta.south);
\node[lbl, anchor=west] at (12.2, 1.0) {(b) learning the operator for inversion};
\node[ex] (gen) at (13.6, 0) {generic operator:\\procedural media};
\node[ex] (scope) at (13.6, -1.45) {scope operator:\\+ media of the scope};
\node[op] (traj) at (13.6, -2.95) {\textbf{trajectory operator}:\\+ states the inversion visits};
\node[box, draw=cOurs!80!black, line width=0.8pt, fill=white] (visit) at (16.75, -1.45)
  {inversions with the\\operator in the loop};
\node[mc] (lab) at (16.75, -2.45) {Monte Carlo labels};
\draw[fwd] (gen) -- node[right, lbl] {fine-tune} (scope);
\draw[fwd] (scope) -- node[right, lbl] {fine-tune} (traj);
\draw[grad] (scope.east) -- (visit.west);
\draw[grad] (visit.south) -- (lab.north);
\draw[grad] (lab.south) |- (traj.east);
\end{tikzpicture}

%% file: sec/3_method.tex
\section{Method}
\label{sec:method}

We reconstruct a heterogeneous participating medium from photographs taken under known directional illumination, with the goal of relighting it from novel views and light directions. 

\subsection{Image Formation and Gradient Decomposition}
\label{sec:method:image}

The unknown is the extinction field $\sigt(\vx)$ in the unit cube, represented on a voxel grid. The albedo $\alpha$ and Henyey--Greenstein phase-function parameter $g$ are known, and the scattering coefficient is $\sigs=\alpha\sigt$. For light propagating along $\wl$, the direct-beam transmittance at $\vx$ is $\Tl(\vx)=e^{-\taul(\vx)}$, where $\taul(\vx)=\int_0^\infty\sigt(\vx-s\wl)\,ds$. A camera ray $\vx_t=\vo+t\vd$ observes
\begin{equation}
  I=\int_0^\infty \Tcam(t)\,\sigs(\vx_t)\big[\Tl(\vx_t)\,p_g(\wl\!\cdot\!\wo)+S(\vx_t,\wo)\big] \,dt,
  \label{eq:render}
\end{equation}
where $\wo=-\vd$, $p_g$ is the phase function, and $\Tcam$ is transmittance along the camera ray. The first term in brackets accounts for light scattered once toward the camera. The source $S$ collects light that has already scattered and is scattered again toward $\wo$; its dependence on extinction requires solving multiple-scattering transport. Its angular integral is given in the supplement. In contrast, transmittance and single scattering are computed by deterministic ray marching.

We optimize the log-extinction parameters $\theta$, with $\sigt=\min(e^\theta,2000)$. Let $\vI(\theta)\in\mathbb R^P$ contain the rendered values of $P$ rays and $\vr=\vI-\vy$ their residuals. The image Jacobian splits according to the three ways in which extinction enters \cref{eq:render}:
\begin{equation}
  \mJ=\frac{\partial\vI}{\partial\theta}=\Jcam+\Jdir+\Jms.
  \label{eq:decomp}
\end{equation}
$\Jcam$ captures changes in the local scattering coefficient and camera transmittance while holding the bracketed term in \cref{eq:render} fixed; $\Jdir$ captures changes in the incoming direct illumination, while $\Jms$ captures changes in the multiple-scattering field. Frozen-source tomography keeps only $\Jcam$, whereas differentiable path tracing estimates the full gradient $\mJ^\top\vr$ by Monte Carlo. Our method retains $\Jcam$ and $\Jdir$ and learns the expensive $\Jms$ term.

\subsection{Neural Multiple-Scattering Operator}
\label{sec:method:operator}

\paragraph{Operator and representation.}
We learn the multiple-scattering source rather than the rendered image. A neural operator $G_\phi$ takes the medium and illumination as input and predicts angular coefficients of the source field throughout the volume. A single predicted field is shared across all camera rays and views; only its evaluation direction changes. The input includes extinction, albedo, light transmittance, and analytic single- and forward-scattering features. The backbone is a 3D U-Net~\cite{cicek2016unet3d}, conditioned on the phase-function parameter $g$ and light direction $\wl$ through FiLM~\cite{perez2018film}.

We represent the angular dependence using a basis $\mathcal B$ containing real spherical harmonics (SH) through degree $4$ (25 functions) and 12 additional light-aligned zonal harmonics of degrees $5$--$16$. Under Henyey--Greenstein scattering, harmonics of degree $\ell$ have eigenvalue $g^\ell$ (Funk--Hecke, as in MRBNN~\cite{liang2026mrbnn}), so the source is reconstructed from angular moments $c_b$ of the radiance that has scattered at least once:
\begin{equation}
  S(\vx,\vw)=\sum_{b\in\mathcal B}g^{\ell(b)}\,c_b(\vx)\,b(\vw).
  \label{eq:funkhecke}
\end{equation}
Here $\ell(b)$ denotes the degree of basis function $b$. The zonal terms resolve the forward-scattering lobe without the 264 additional coefficients required by full SH through degree $16$. The network predicts a light-aligned zonal component and a world-frame SH residual, which are mapped to the coefficients in \cref{eq:funkhecke}; the exact basis and mapping are in the supplement.

\begin{figure}[t]
  \centering
  \includegraphics[width=\linewidth]{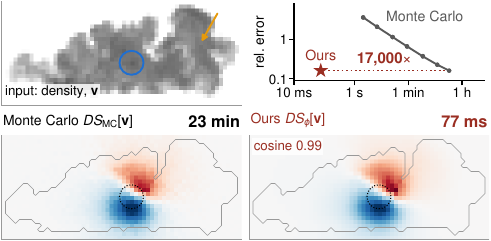}
  \caption{\textbf{Accurate multiple-scattering derivatives at low cost} -- Response of the multiple-scattering source
    (angular mean; shared color scale, red: increase, blue: decrease) to a local density increase $\vv$ (circle;
    vertical slice of the cloud of \cref{fig:teaser}; arrow: input light). One Monte Carlo central difference
    (\dtRefPhotons{} photons per evaluation) takes \dtMatch{} and reaches our error; ours takes \dtOurs{} (each on one
    \dtGPU, offline training excluded): about \dtMatchRatio{} times faster for this directional derivative (not
    inversion gradients or PRB). Cosine and errors are $\sigs$-weighted against a reference of two independent estimates
    (\dtRef{} in total; the first is shown). Top right: error over time (Monte Carlo: noise; ours: bias).}
  \label{fig:opderiv}
\end{figure}

\paragraph{Learning and differentiating the field.}
The generic operator is trained on $1.2\times10^5$ procedural media, covering optical depths $1$--$100$, albedos $0.8$--$0.999$, $g\in[0,0.9]$, and varying light directions. Monte Carlo transport supplies noisy field labels. We write $\Sop(\sigt,\taul)$ for the predicted source, with the other physical parameters understood. Since $\taul$ itself depends on $\sigt$, automatic differentiation through the light sweep, operator, and camera ray marcher supplies $\Jmsphi^\top\vr$ without explicitly forming a Jacobian. The transmittance and single-scattering derivatives remain physically computed. \Cref{fig:opderiv} compares the resulting directional derivative with Monte Carlo in accuracy and online cost.

\subsection{Differentiable Inverse Rendering}
\label{sec:method:inverse}

\begin{figure*}[t]
  \centering
  \includegraphics[width=\textwidth]{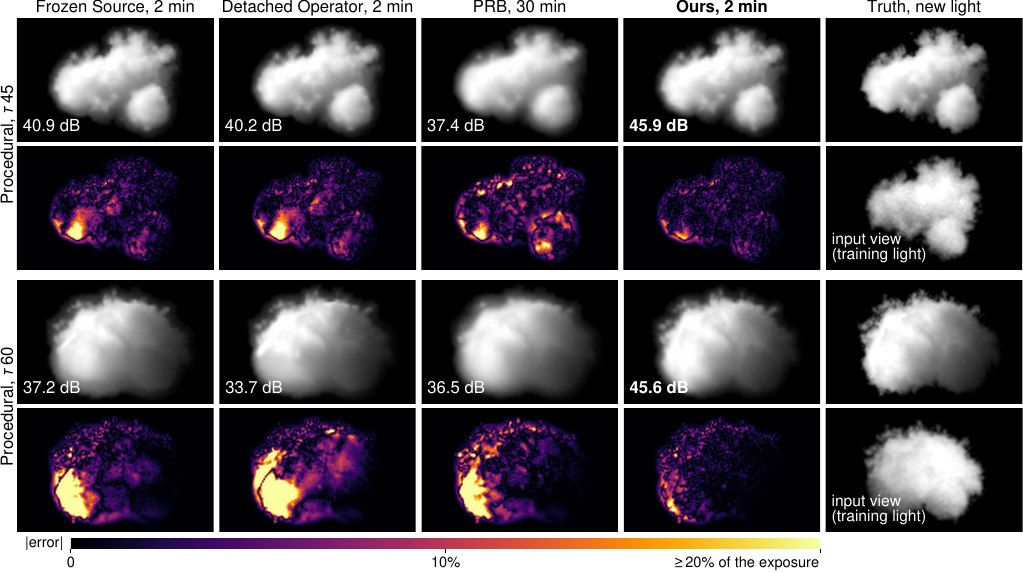}
  \caption{\textbf{Relighting two additional clouds} -- The media of the main evaluation at the 25th and 75th percentile of our
    gain over \FS
    (procedural cumulus, maximum optical depth 45 and 60), shown as in \cref{fig:teaser} under their most back-lit new
    light (errors saturate at 20\% of the exposure); without the light-side derivative large parts of the clouds are relit
    wrongly.}
  \label{fig:gallery}
\end{figure*}

We reconstruct extinction at $64^3$ resolution, with known albedo, phase function, light, and cameras. At each iteration, we sample training-view rays $\mathcal R$ and minimize their squared image residuals, $\mathcal L=|\mathcal R|^{-1}\sum_{p\in\mathcal R}(I_p-y_p)^2$. The light sweep gives $\taul$ and $\Tl$; the operator predicts $\Sop$; and ray marching evaluates \cref{eq:render}. We use the corrected forward source $S=\Sop(\sigt,\taul)+\Delta$, where $\Delta$ is held constant during differentiation. The resulting update uses
\begin{equation}
  \nabla_\theta\mathcal L=\frac{2}{|\mathcal R|}\big(\Jcam+\Jdir+\Jmsphi\big)^\top\vr,
  \label{eq:grad}
\end{equation}
with $\vr$ restricted here to the sampled rays. Adam updates $\theta$ using this gradient.

\paragraph{Monte Carlo correction.}
Every 30 iterations, a new Monte Carlo estimate $\Smc$ updates $\Delta$ by an exponential moving average with weight $\beta=0.25$ of the stop-gradient difference $\Smc-\Sop$. This anchors the forward values without replacing the operator's learned Jacobian. The correction is initialized from the current Monte Carlo--operator difference at the first refresh and after each coarse-to-fine resolution change; otherwise it uses the moving average. It can lag as density changes, making operator accuracy along the trajectory important.

\paragraph{Gradient baselines and output.}
\FS uses periodically refreshed Monte Carlo source fields but holds the entire light-side term fixed during differentiation, retaining only $\Jcam$. \DO uses the same anchored source values as ours but likewise retains only $\Jcam$; our method additionally differentiates the direct beam and the operator-predicted multiple scattering. Every 25 iterations we evaluate four held-out views and retain the best checkpoint. The inversion uses a coarse-to-fine warm start and runs at approximately 17 iterations per second at $64^3$. The full procedure and baseline protocols are in the supplement and \cref{sec:exp:setup}.

\subsection{Learning the Operator for Inversion}
\label{sec:method:oplearn}

During inversion, the operator encounters density states that may lie outside its training distribution. Inversion can create transient, speckled, or sharply peaked density fields; derivative errors on these states redirect subsequent updates, which in turn change the states seen by the operator. We therefore train on the distribution induced by inversion itself (\cref{fig:overview}b).

Training has three stages. \emph{Generic pretraining} covers a broad range of media and illumination (\cref{sec:method:operator}). \emph{Scope fine-tuning} specializes the operator to strongly scattering media using 4000 procedural and Cloudy~\cite{leonard2025ltdps} training volumes (optical depths $30$--$120$, albedos $0.99$--$0.9999$, and $g\in[0.6,0.9]$), mixed equally with generic data. \emph{Trajectory fine-tuning} uses the scope operator to reconstruct separate training media, collects the density states visited during these inversions, labels their scattering fields with Monte Carlo, and fine-tunes on these states together with the earlier data. This follows the data-aggregation principle of DAgger~\cite{ross2011dagger}.

We supervise scattering-field values rather than Jacobians. Sampling states along an inversion trajectory also constrains the predicted field's changes along the displacements that the optimizer actually makes. The resulting improvements in derivative alignment and optimization stability are measured in \cref{sec:exp:trajectory}. Each fine-tuning stage uses $4\times10^4$ training steps; equal-steps controls isolate the effect of the added data from additional training time. Implementation and data-collection details are in the supplement; the reconstruction loop is summarized in \cref{alg:ours}.

\begin{algorithm}[t]
  \small
  \caption{\textbf{Operator training and inverse rendering}}
  \label{alg:ours}
  \KwIn{Generic operator $\phi_g$, generic/scope data $\mathcal D_g,\mathcal D_s$, observations $\vy$}
  \textbf{Train the operator}\;
  $\phi_s\gets\operatorname{FineTune}(\phi_g,\mathcal D_g\cup\mathcal D_s)$\;
  $\mathcal T\gets\operatorname{VisitedStates}(\operatorname{Invert}(\vy_m,\phi_s))$
  over separate training media $m$\;
  $\mathcal D_t\gets\{(\sigma,\Smc(\sigma)):\sigma\in\mathcal T\text{ accepted}\}$\;
  $\phi_t\gets\operatorname{FineTune}(\phi_s,\mathcal D_g\cup\mathcal D_s\cup\mathcal D_t)$\;
  \BlankLine
  \textbf{Invert with $G_{\phi}$ ($\phi_t$ at test time)}\;
  $\theta\gets\operatorname{WarmStart}(\vy)$; $\theta^*\gets\theta$; $\ell^*\gets\infty$\;
  \For{step $k=1,2,\ldots$ until the time budget}{
    \lIf{grid resolution changes}{Upsample $\theta$; reset Adam state and anchor}
    $\sigt\gets\min(e^\theta,2000)$; $\taul\gets\operatorname{Sweep}(\sigt)$\;
    \lIf{initial, 30-step, or grid-switch refresh}{UpdateAnchor$(\Delta,\sigt,\taul,\phi)$}
    $S\gets\Sop(\sigt,\taul)+\sg[\Delta]$; sample rays $\mathcal R$\;
    $\vr\gets\operatorname{Render}(\mathcal R;\sigt,S)-\vy_{\mathcal R}$\;
    $\theta\gets\operatorname{AdamStep}(\theta,\nabla_\theta\mathcal L)$ from \cref{eq:grad}\;
    \lIf{$k\bmod25=0$}{Keep $\theta^*\gets\theta$ if held-out loss improves}
  }
  \KwRet{$\theta^*$}
\end{algorithm}

%% file: sec/4_experiments.tex
\section{Experiments}
\label{sec:exp}

\subsection{Experimental Setup}
\label{sec:exp:setup}

\paragraph{Data and task.}
We evaluate on held-out, optically thick clouds with maximum column optical depth 40--100, albedo 0.99--0.999, and $g\in[0.7,0.9]$. Half are procedural cumulus and half come from Cloudy~\cite{leonard2025ltdps}; none are used for training or development. Our \emph{main evaluation} contains 12 media rendered by Mitsuba~3~\cite{jakob2022mitsuba3} at $128^3$ resolution, while all methods reconstruct extinction at $64^3$. The observations therefore differ from every method's reconstruction model. The protocol and success criteria were fixed before generating these test media (supplement). A separate 12-medium \emph{controlled set}, rendered at $64^3$ by our path tracer, supports comparisons and ablations; the supplement also reports a broader optical-depth study.

Each medium is observed in 32 views of $128^2$ pixels under one directional light. We evaluate reconstruction quality by rendering three novel light directions from two unseen views. Our primary metric is relighting PSNR against reference images, corrected for the Monte Carlo noise in both images (supplement). We also report voxel-wise extinction error and projected-optical-depth error.

\paragraph{Baselines and timing.}
\FS~\cite{levis2015airborne} uses periodically refreshed Monte Carlo source fields~\cite{holodovsky2016insitu} but retains only the camera-side derivative. The \DO uses our corrected operator source with the same detached derivative, isolating the contribution of differentiating through the operator. PRB~\cite{vicini2021prb} estimates the full gradient in Mitsuba~3. All methods use the same density parameterization, optimizer, 10-s warm start, and visual hull. We select the best checkpoint using four held-out views, except that PRB returns its current iterate. \FS and PRB use the best of their evaluated configurations for each medium and budget, including two smoothness priors; ours and the \DO use one fixed configuration. PRB uses no Russian roulette on the main evaluation and Mitsuba's default roulette on the controlled set; budget-specific tuning is detailed in the supplement.

We compare equal wall-clock budgets of 30\,s and 2, 8, and 16\,min on one NVIDIA B200, including warm start and checkpoint selection. PRB is additionally run to 30\,min. After 2\,min, \FS performs 1.7--1.8 times as many iterations as ours, while the tested PRB configurations complete 0.02--0.24 times as many. Unless stated otherwise, we report medians across media and bootstrap 95\% intervals for median paired differences.

\subsection{Reconstruction and Relighting}
\label{sec:exp:main}

\begin{table}[t]
  \centering\footnotesize\setlength{\tabcolsep}{2.2pt}
  \resizebox{\linewidth}{!}{\input{tables/main_v2.tex}}
  \caption{\textbf{Relighting at equal time} -- Median relighting PSNR [dB] over the 12 media of the main evaluation
    after each budget (top) and median per-medium gain of ours (bottom; positive on all 12 media in every cell). Ours
    after 2\,min leads PRB after 30\,min (33.4\,dB) on all 12 media, by 9.0\,dB [8.1, 9.9]. Jacobian: the terms of
    \cref{eq:decomp} in the gradient. \FS: best of two configurations per medium and budget. PRB: best of four at 30\,s and 2\,min (two added in a subsequent evaluation), and of two at longer budgets.}
  \label{tab:main}
\end{table}

\paragraph{Relighting under new illumination.}
Differentiating through the learned multiple-scattering source improves relighting on every main-evaluation medium at every tested budget (\cref{tab:main}). After 2\,min, ours reaches a median of 42.5\,dB, compared with 35.1\,dB for \FS, 32.8\,dB for the \DO, and 29.8\,dB for PRB. The median paired gain over \FS is 6.8\,dB [5.5, 10.7] (12/12 media), and remains 4.0\,dB after 16\,min. PRB reaches 33.4\,dB after 30\,min; ours after 2\,min still leads on all 12 media, by a median paired 9.0\,dB [8.1, 9.9]. The controlled set gives the same result with the scope-trained operator: at 2\,min ours gains 6.6\,dB over \FS, and it exceeds PRB after 30\,min by 12.0\,dB with roulette (10.7\,dB without roulette, post hoc), each on all 12 media.

The relighting examples in \cref{fig:teaser,fig:gallery} expose the weakness of camera-side gradients. Camera-side gradients can fit the observed images while placing extinction incorrectly inside the cloud; the error is revealed by a new light. With the learned light-side derivative, the reconstruction retains the structure needed for relighting.

\begin{figure*}[t]
  \centering
  \includegraphics[width=\textwidth]{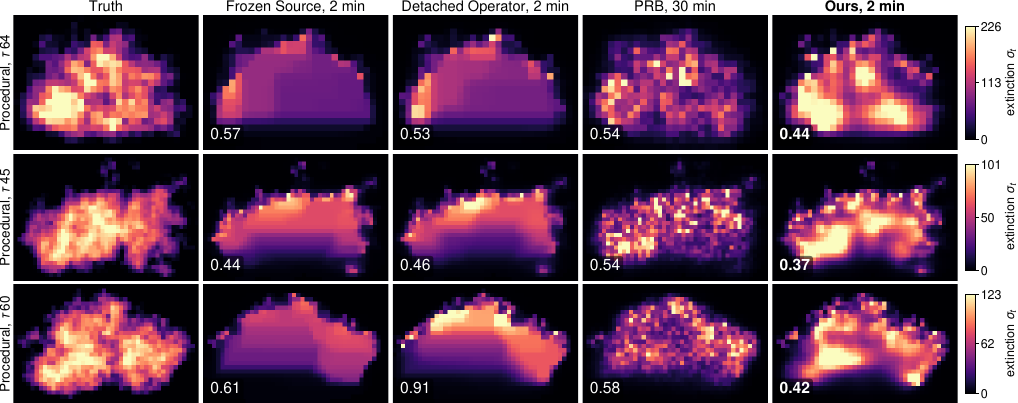}
  \caption{\textbf{What is reconstructed} -- Vertical density slices: a ramp without the light-side derivative, speckle
    for PRB, clumps for ours (number: relative density error over the grid, lowest in bold). Top: near the median margin
    among the 6 of 12 media where ours has the lowest density error of the methods shown (PRB after 30\,min is lower on
    five Cloudy ones); below: the media of \cref{fig:gallery}.}
  \label{fig:slices}
\end{figure*}

\paragraph{Recovered extinction.}
The density slices show corresponding reconstruction errors (\cref{fig:slices}). Without the multiple-scattering derivative, extinction tends toward smooth interior ramps; PRB's noisy updates produce speckled density, whereas ours recovers more of the underlying clumps. The gap is larger in projected optical depth: after 2\,min its median relative error is 0.20 for ours, versus 0.42--0.68 for the two \FS priors, 0.53 for the \DO, and 0.31--0.36 for PRB after 30\,min (\cref{fig:supp:od}). The voxel-wise extinction errors are closer (ours 0.55; other methods 0.55--1.29). Ours has the lowest voxel-wise error on 6 of 12 main-evaluation media among the shown methods; PRB is lower on five Cloudy media and \FS on one.

\subsection{The Contribution of the Learned Derivative}
\label{sec:exp:derivative}

\paragraph{An accurate nonlocal response at low cost.}
For the cloud shown in \cref{fig:teaser}, the predicted multiple-scattering source differs from Monte Carlo by \opfFigVal{} ($\sigs$-weighted relative field error; \cref{tab:supp:opfield}). Along a localized density increase, its directional derivative agrees with the Monte Carlo central-difference reference (cosine \opfFigCos{}, relative error \dtOursErr); \cref{fig:opderiv} shows the response extending well beyond the perturbed region. Outside two standard deviations of the perturbation, \opfFigFarShare{} of the angular-mean response's squared norm lies, with cosine \opfFigFarCos{}. Evaluating this directional derivative takes \dtOurs{} for the operator versus \dtMatch{} for Monte Carlo at comparable reference-relative error. Across the 12 main-evaluation media, source-field errors are \opfValRange{} and derivative cosines \opfCosRange{} (post hoc; supplement). In rendered-image responses, the frozen-source derivative captures only \opfFrozenRange{} of the true change's norm (median \opfFrozenMed), while ours preserves its direction (cosine \opfImgCosRange{}; \cref{tab:supp:opfield}).

\paragraph{The multiple-scattering term dominates.}
For four density perturbations on development clouds, $\|\Jms\vv\|$ is 2.9--4.8 times $\|\Jcam\vv\|$, while $\|\Jdir\vv\|$ is only 0.24--0.34 times $\|\Jcam\vv\|$ (the decomposition uses the generic operator; supplement). The direct-light response opposes the multiple-scattering response: upstream extinction blocks the direct beam but also redirects scattered light. As optical depth increases from 5 to 60, the frozen-source gradient's cosine with the full gradient drops from 0.95 to 0.58.

\paragraph{The derivative, not just the field values.}
The \DO shares our corrected source values but omits both $\Jdir$ and $\Jmsphi$. It trails ours by 10.2\,dB after 2\,min on the main evaluation (12/12 media; \cref{tab:main}). On the controlled set, detached operator values perform almost identically to the Monte Carlo values used by \FS (median difference $-0.03$\,dB at 2\,min, post hoc), while enabling the learned derivative improves relighting by 7.1\,dB. Replacing the learned operator with a classical P1 diffusion model under the same correction yields 26--36\,dB on eight development media, 13.8\,dB below our operator and below \FS after 2 and 8\,min.

\subsection{Learning on Inversion Trajectories}
\label{sec:exp:trajectory}

\begin{table*}[t]
  \centering\small\setlength{\tabcolsep}{4.2pt}
  \input{tables/oplearn_v2.tex}
  \caption{\textbf{Operators for inversion} -- Operator quality (development): field error on generic media and on 95
    held-back visited states; cosine of the derivative with Monte Carlo's along the inversion's next step from drifted
    states. Inversion performance: drift (median density error of the current iterate after 16\,min, with selection) and
    median relighting PSNR [dB] of the final iterate (no selection) and of the selected checkpoint. +: equal-steps
    controls without the new data.}
  \label{tab:oplearn}
\end{table*}

\paragraph{Accuracy where the optimizer goes.}
The operators in \cref{tab:oplearn} have generic-media field errors within 7\% of each other, yet their behavior inside inversion differs sharply. The trajectory operator and the equal-steps scope+ control have nearly identical generic errors (0.0216 and 0.0214), but the trajectory operator has lower error on held-back inversion states (0.0446 versus 0.0597). On drifted states, its derivative along the next optimization step aligns much better with Monte Carlo (cosine 0.51 versus 0.13--0.18 for the other operators; The equal-steps controls show that this improvement comes from the visited-state data rather than additional training.

\begin{figure}[t]
  \centering
  \includegraphics[width=\linewidth]{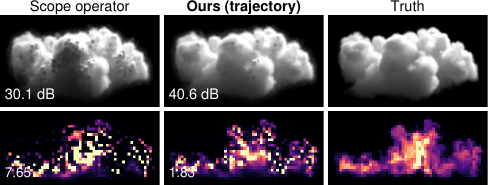}
  \caption{\textbf{Trajectory training reduces optimization drift} -- The medium of \cref{fig:teaser} after 16\,min on
    all 32 views, final iterate without held-out selection: relit (top, PSNR) and density slice (bottom, density error).
    With the scope operator the iterate runs into density spikes; with the trajectory operator it stays far closer to
    the truth.}
  \label{fig:drift}
\end{figure}
\paragraph{Optimization stability.}
Trajectory training also reduces final-iterate drift. After 16\,min, its current-iterate density error on the main evaluation is lower than that of scope and scope+ on all 12 media (median per-medium error ratios 0.28 and 0.41). Without held-out checkpoint selection, the scope and scope+ operators diverge on 11 of 12 media, versus 4 of 12 for trajectory. \Cref{fig:drift} shows one such case: the scope operator accumulates density spikes, while trajectory training preserves a more faithful reconstruction.

At the final iterate, without selection, the trajectory operator's median paired gain over scope is 4.5\,dB after 8\,min (11/12 media) and 8.7\,dB after 16\,min (12/12), and over scope+ 5.0 and 8.0\,dB. With held-out checkpoint selection, the improvement is smaller: 0.70\,dB averaged over main-evaluation media and budgets ($t$-interval [0.26, 1.14]; controlled set: 0.44\,dB). Trajectory training primarily stabilizes inversion, while checkpoint selection remains useful for the residual drift.

\subsection{Ablations}
\label{sec:exp:ablations}

The multiple-scattering derivative carries the improvement. On the controlled set, removing $\Jdir$ changes relighting by only 0.1--0.3\,dB in favor of the variant without it, with every interval including zero. Dropping $\Jmsphi$ as well, as in the \DO, reduces relighting by 7.1\,dB after 2\,min (\cref{tab:supp:ablation}). Removing the Monte Carlo correction costs 0.8--0.9\,dB from 2\,min onward, though it gains 0.5\,dB at 30\,s because no correction refresh is needed. \paragraph{Dependence on optical thickness.}
The benefit is strongest when multiple scattering dominates. On moderate development media ($\tau=7.6$--24), ours with the scope operator and \FS are tied after 2\,min (median $-0.01$\,dB). Within the controlled set, the 2-min median gain is 7.8\,dB on eight media with $\tau\geq50$, versus 1.5\,dB on the other four ($\tau<50$; post hoc). This agrees with the increasingly poor frozen-source gradient alignment at larger optical depths (supplement, Sec.~\ref{sec:supp:regime}).

Architecture and zonal-basis ablations are reported in the supplement.

%% file: tables/main_v2.tex
\begin{tabular}{@{}llcccc@{}}
\toprule
Method & Jacobian & 30\,s & 2\,min & 8\,min & 16\,min \\
\midrule
\multicolumn{6}{@{}l}{\emph{Median PSNR [dB]}} \\
\FS~\cite{levis2015airborne} & $\Jcam$ & 31.3 & 35.1 & 40.1 & 40.0 \\
\DO & $\Jcam$ & 32.2 & 32.8 & 38.6 & 39.9 \\
PRB~\cite{vicini2021prb} & MC est.\ of $\mJ$ & 29.3 & 29.8 & 31.2 & 32.3 \\
\textbf{Ours} & $\Jcam{+}\Jdir{+}\Jmsphi$ & \textbf{37.3} & \textbf{42.5} & \textbf{42.9} & \textbf{42.9} \\
\midrule
\multicolumn{6}{@{}l}{\emph{Median paired gain of Ours [dB]}} \\
\multicolumn{2}{@{}l}{over \FS} & $+5.1$ & $+6.8$ & $+4.4$ & $+4.0$ \\
\multicolumn{2}{@{}l}{over the \DO} & $+5.2$ & $+10.2$ & $+5.9$ & $+5.1$ \\
\bottomrule
\end{tabular}

%% file: tables/oplearn_v2.tex
\begin{tabular}{@{}lccccccccc@{}}
\toprule
 & \multicolumn{3}{c}{operator quality (development)} & \multicolumn{6}{c}{inversion performance} \\
\cmidrule(lr){2-4}\cmidrule(l){5-10}
 & \multicolumn{2}{c}{field error $\downarrow$} & cosine $\uparrow$ & \multicolumn{2}{c}{drift $\downarrow$}
 & \multicolumn{2}{c}{final iterate, main $\uparrow$} & \multicolumn{2}{c}{selected, 16\,min $\uparrow$} \\
\cmidrule(lr){2-3}\cmidrule(lr){4-4}\cmidrule(lr){5-6}\cmidrule(lr){7-8}\cmidrule(l){9-10}
Operator & generic & visited & next step & controlled & main & 8\,min & 16\,min & controlled & main \\
\midrule
generic                   & 0.0228 & 0.0725 & 0.15 & 6.42 & 7.54 & 35.2 & 30.8 & 42.8 & 41.3 \\
generic+ (equal steps)    & 0.0229 & 0.0693 & 0.18 & 6.74 & 7.80 & 38.2 & 30.0 & 43.2 & 41.8 \\
scope                     & 0.0222 & 0.0630 & 0.13 & 4.09 & 7.66 & 40.0 & 30.5 & 43.7 & 42.4 \\
scope+ (equal steps)      & 0.0214 & 0.0597 & 0.15 & 3.74 & 5.54 & 40.3 & 30.8 & 43.9 & 42.6 \\
\textbf{trajectory (ours)} & 0.0216 & \textbf{0.0446} & \textbf{0.51} & \textbf{1.77} & \textbf{1.49} & \textbf{43.6}
 & \textbf{43.0} & \textbf{44.7} & \textbf{42.9} \\
\bottomrule
\end{tabular}

%% file: sec/5_conclusion.tex


\section{Conclusion and Future Directions}
\label{sec:conclusion}
A neural operator can supply the costly multiple-scattering
derivative needed for inverse volume rendering. Differentiating
its predicted scattering field avoids repeated Monte Carlo
gradient estimation while retaining the nonlocal response
omitted by frozen-source tomography. Training the operator on inversion-visited
states further improves the reliability of its derivatives and
substantially reduces optimization drift. The resulting method improves reconstruction and relighting of optically thick clouds at fixed computational budgets.

Next steps include smoke and other strongly scattering media, jointly recovering illumination and material parameters, and reconstruction from captured images. The same principle of learning a physical field whose derivatives drive inference may extend to other inverse problems with expensive gradients.

%% file: sec/X_suppl.tex

\clearpage
\maketitlesupplementary
\zlabel{supp:start}

\appendix
\crefname{appendix}{Sec.}{Secs.}
\Crefname{appendix}{Section}{Sections}
\crefname{subappendix}{Sec.}{Secs.}
\Crefname{subappendix}{Section}{Sections}
\setcounter{figure}{0}\renewcommand{\thefigure}{S\arabic{figure}}      
\setcounter{table}{0}\renewcommand{\thetable}{S\arabic{table}}
\setcounter{equation}{0}\renewcommand{\theequation}{S\arabic{equation}}
\setcounter{algocf}{0}\renewcommand{\thealgocf}{S\arabic{algocf}}


\section{Method Details}
\label{sec:supp:method}

\subsection{Renderer and Angular Representation}
\label{sec:supp:render}

\paragraph{Ray marching.}
Grids hold values at the voxel centres of the unit cube and are interpolated trilinearly, constant between the
outermost centres and the faces and zero outside the cube; the ray marcher, the Monte Carlo kernels and Mitsuba~3 use
the same convention.
A camera ray is marched from its entry into the cube with a fixed step $h=1/(2N)$, two samples per voxel at $N=64$;
$\sigt$, $\taul$ and $S$ are read at the step midpoints and held constant over the step, so step $j$ contributes
$T_j\,(1-e^{-\sigt h})\,\alpha\,Q_j$, with $T_j$ the transmittance to the start of the step and
$Q_j=\Tl\,p_g(\wl\!\cdot\!\wo)+S$: \cref{eq:render} integrated exactly for piecewise-constant coefficients.
The optical depth to the light is computed at the voxel centres by marching towards the light with the same step;
camera samples interpolate $\taul$ and exponentiate, since interpolating $\Tl$ biases single scattering by $+0.1$ to
$+0.8\%$.
For full multiple scattering our path tracer agrees with pbrt-v3 (mean ratio 0.9971--1.0000 at $g=0$, 0.5 and 0.85)
and with Mitsuba~3.9.1 to within Mitsuba's own seed-to-seed noise.

\paragraph{Angular basis.}
The source of \cref{eq:render},
$S(\vx,\vw)=\int_{\mathbb{S}^2}p_g(\vw\!\cdot\!\vw')\,\Lms(\vx,\vw')\,d\vw'$ with $\Lms$ the radiance that has
scattered at least once, is evaluated from 37 nodal coefficient grids: real SH to degree 4 (orthonormal, without the Condon--Shortley phase)
and the zonal harmonics of degrees 5--16 about $\wl$, with the factors $g^\ell$ applied.
Multiple-scattering images of a development cumulus ($\tau=20$, albedo 0.99, $g=0.85$) rendered from tallied moments
under a top, a front and a back light reach 32.7, 15.6 and 35.9\,dB with SH to degree 4 alone and 36.7, 33.1 and
41.3\,dB with the zonal extension, the noise floor of the 4096-spp references (34.3, 31.5 and 38.3\,dB between two
independent references): the forward cascade of a front light is what truncated SH miss.
At $g=0.9$, front-lit, zonal degrees to 16 give 19.6--23.4\,dB and to 32 give 31.2--33.5\,dB, so more strongly
forward-scattering media would need a higher zonal degree.
At the true density of five development media, this forward model relights under the evaluation lights at
41.9--51.1\,dB (single scattering alone: 16.2--25.9\,dB).

\paragraph{Cell averages and nodal values.}
Monte Carlo tallies cell averages, whereas rendering interpolates nodal values.
The operator therefore predicts nodal values $\hat{\mathbf{c}}$ and is trained through their cell averages
$\Mcal\hat{\mathbf{c}}$, $\Mcal=\mathbf{M}^{\otimes3}$, so noisy labels are never deconvolved; per axis $\mathbf{M}$
has interior rows $(\frac18,\frac34,\frac18)$ and boundary rows $(\frac78,\frac18)$.
Monte Carlo fields entering the inversion are converted with $\Mcal^{-1}$; rendering tallied cell averages as nodal
values instead would bias multiple scattering by $-0.6$ to $-1.1\%$.

\subsection{Operator}
\label{sec:supp:operator}

\paragraph{Inputs and network.}
The 14 input channels per voxel are $\log(1+\sigt/10)$, $-\log(1-\alpha)/5$, $\Tl$, the first-order forward-lobe
amplitude $\Tl\tau_s$ with $\tau_s(\vx)=\int_0^\infty\sigs(\vx-s\wl)\,ds$, $\log(1+\sigs\Tl/10)$ and, for
$\ell\in\{0,1,2,3,4,6,8,12,16\}$, the zonal moments of the scattered radiance in the small-angle approximation,
\begin{equation}
  z^{\mathrm{fwd}}_\ell=\sqrt{\tfrac{2\ell+1}{4\pi}}\;\Tl\,\big(e^{\tau_s g^\ell}-1\big),
  \label{eq:fwd}
\end{equation}
which reuse the sweep's $\taul$ inside the inversion ($\tau_s=\alpha\taul$ for a constant albedo).
$(g,\wl)$ passes through a two-layer MLP to a 128-dimensional embedding.
The 3D U-Net has pre-activation residual blocks (GroupNorm, SiLU, $3^3$ convolutions), two per level at widths 32,
64, 128 and 256 and two at the bottleneck, strided downsampling and transposed-convolution upsampling with concatenated
skips; FiLM, $x\mapsto x(1+\gamma)+\beta$ with zero-initialised $(\gamma,\beta)$, follows the first convolution of
every block.
The head outputs 42 channels, $z_0,\dots,z_{16}$ and the 25 residuals $u_{\ell m}$, which are multiplied by $g^\ell$
and mapped to the 37 coefficients of $\mathcal{B}$ by
\begin{equation}
  c_{Y_{\ell m}}=\sqrt{\tfrac{4\pi}{2\ell+1}}\,Y_{\ell m}(\wl)\,z_\ell+u_{\ell m},\qquad c_{Z_\ell}=z_\ell ;
  \label{eq:lightframe}
\end{equation}
the network has 19.4\,M parameters, and the network alone takes about 13\,ms per forward pass at $64^3$ in bfloat16.

\paragraph{Generic training.}
Procedural $64^3$ media (cumulus 40\%, fBm 25\%, plume 25\%, blob 10\%; maximum column optical depth log-uniform in
$[1,100]$, $1-\alpha$ log-uniform in $[0.001,0.2]$, spatially varying in 25\% of the media, $g$ uniform in $[0,0.9]$,
light uniform on the sphere) are generated on line: at every step one new medium is tallied with $2\times10^6$ photons
and replaces the oldest of a buffer of 1024, from which batches of 4 are drawn.
The loss is the relative squared error of $\Mcal\hat{\mathbf{c}}$ against the tallied cell averages on the medium
dilated by one voxel.
AdamW~\cite{loshchilov2019adamw} runs at learning rate $10^{-3}$ after 500 warm-up steps with cosine decay over
$1.2\times10^5$ steps, weight decay $10^{-4}$, gradient clipping at 1 and bfloat16 mixed precision; the selected
checkpoint, the last, has a field error of 0.0228 on 200 validation media (geometric mean of the relative $L_2$ error).
Training took 5.3\,h on one NVIDIA B200.
\Cref{tab:supp:opabl} changes one design choice at a time.

\begin{table}[t]
  \centering\scriptsize\setlength{\tabcolsep}{3pt}
  \input{tables/op_ablation.tex}
  \caption{\textbf{Operator design} -- One change at a time against the reference configuration (30k training
    steps; 200 procedural validation media): field error and image PSNR [dB] of single plus multiple scattering and of
    multiple scattering alone.}
  \label{tab:supp:opabl}
\end{table}

\paragraph{Forward accuracy on held-out media.}
On 300 held-out procedural and Cloudy media (the forward test of the regime set), images rendered with the operator's field are
15.9\,dB more accurate in the median than with a Monte Carlo field computed in the same time ($6\times10^5$ photons;
better on 100\%) and 1.2\,dB more accurate than with one computed in about 32 times the time ($2\times10^7$ photons;
better on 66\%), against a $2\times10^8$-photon reference.
Outside the training distribution, on 100 media from The Well~\cite{ohana2024well} (MHD and supernova), it stays ahead
of the $6\times10^5$-photon field (by 9.6 and 12.2\,dB; there it took 1.6--3.2 times the operator's time) but behind
the $2\times10^7$-photon one ($-5.1$ and $-2.7$\,dB).

\subsection{Learning the Operator for Inversion}
\label{sec:supp:oplearn_data}

The training and inversion procedures are summarized in \cref{alg:ours} in the main paper; implementation schedules and data selection are specified below.

\paragraph{Scope data.}
4000 training and 100 validation media in which multiple scattering dominates: maximum column optical depth
log-uniform in $[30,120]$, albedo in $[0.99,0.9999]$ ($1-\alpha$ log-uniform, spatially varying in 25\% of the media),
$g$ uniform in $[0.6,0.9]$, light direction uniform; procedural cumulus 35\%, fBm 15\%, plume 10\%, blob 5\% and
Cloudy training clouds 35\%.
Labels: $2\times10^7$ photons per training medium, $2\times10^8$ per validation medium.

\paragraph{Trajectory data.}
A separate set of 64 training-side media is reconstructed by inversions with the scope operator in the loop
(18000 iterations, coarse to fine at 300 and 900 iterations, observations at 1024 samples per pixel, so that the
trajectories also visit noise-driven states); the density is stored every 750 iterations.
Every stored state with density error at most 3 is labelled by Monte Carlo ($2\times10^7$ photons): 896 states of 56
training media (448 further, fully diverged states are skipped); the remaining 8 media give 95 held-back states,
labelled with $10^8$ photons, on which the operators are compared.

\paragraph{Fine-tuning.}
Every fine-tuning stage starts from the previous operator and runs $4\times10^4$ steps (batch 4, AdamW, learning
rate $3\times10^{-4}$ after 500 warm-up steps, weight decay $10^{-4}$, gradient clipping at 1, mixed precision), with
selection on the scope validation set every 2000 steps (\cref{tab:supp:stages}).
The equal-steps controls start from the same operator and run the same schedule without the new data.

\begin{table}[t]
  \centering\footnotesize\setlength{\tabcolsep}{3pt}
  \begin{tabular}{@{}llll@{}}
    \toprule
    Operator & starts from & training data & steps \\
    \midrule
    generic    & --       & generic (on line)                   & $1.2\times10^5$ \\
    generic+   & generic  & generic                             & $4\times10^4$ \\
    scope      & generic  & generic $\cup$ scope                & $4\times10^4$ \\
    scope+     & scope    & generic $\cup$ scope                & $4\times10^4$ \\
    trajectory & scope    & generic $\cup$ scope $\cup$ visited & $4\times10^4$ \\
    \bottomrule
  \end{tabular}
  \caption{\textbf{Training stages} -- The equal-steps controls of the scope and trajectory stages are generic+ and
    scope+. Scope: 4000 media; visited: 896 states of inversions with the scope operator.}
  \label{tab:supp:stages}
\end{table}

\subsection{Monte Carlo}
\label{sec:supp:mc}

\paragraph{Light-side fields.}
The moments $c_b$ are estimated by a CUDA photon tracer: photons enter through the lit faces in proportion to their
projected area and are transported by delta tracking with a global majorant, analog absorption and
Henyey--Greenstein sampling; every path segment after the first real scattering is tallied with the track-length
estimator into every voxel it crosses, in double precision, which estimates the cell averages of $\int\Lms b\,d\vw$
without bias.
The uncollided tally reproduces the exact sweep and the absorbed energy balances, both to the level of the noise.
On one B200 the tracer runs at $1.1$--$3.2\times10^8$ photons per second on a $64^3$ cumulus of maximum optical depth
10--50.

\paragraph{Anchor schedule.}
In the inversion a refresh uses $3\times10^6$ photons and a new seed (about 0.11\,s), every 30 iterations and right
after every coarse-to-fine switch, where the average restarts with $\Delta=\Smc-\Sop$; \FS updates its source field
$S\leftarrow(1-\beta)S+\beta\Smc$ on the same schedule.

\paragraph{Fixed point and lag.}
If the iterates converge to $\sigt^\ast$, the exponential average gives
$\mathbb{E}[\Delta]\to\mathbb{E}[\Smc(\sigt^\ast)]-\Sop(\sigt^\ast)$, and the corrected source $S=\Sop+\Delta$ equals the
Monte Carlo one in expectation; the stationarity condition of \cref{eq:grad} differs from the exact one by
$(\Jms-\Jmsphi)^\top\vr$, while that of the frozen source differs by $(\Jdir+\Jms)^\top\vr$.
Away from a fixed point the anchor lags: at $\sigt=\sigt^a+\vdelta$ near the last refresh $\sigt^a$ the source's error is
$(\partial\Sop-\partial S)\,\vdelta$ to first order, with $\partial$ the Jacobians of the sources.
Where the iterate leaves the operator's training distribution, the Jacobian error on the visited states therefore
drives the error of both the gradient and the values, which is what trajectory training targets.

\subsection{Inverse Renderer}
\label{sec:supp:inverse}

\paragraph{Parameterisation and visual hull.}
The density used for rendering, the sweep, the operator, Monte Carlo and evaluation is $\min(e^{\theta},2000)$ (an
optical depth of 31 per voxel at $64^3$), upsampled trilinearly to $64^3$ from the current coarse-to-fine grid and
multiplied by a visual hull~\cite{laurentini1994visual,kutulakos2000theory}: a voxel is kept if it projects onto a
pixel brighter than 0.5\% of the image maximum in every training view that sees it, and the mask is dilated by one
voxel.
The hull and every other shared setting apply to every method, PRB included.

\paragraph{Warm start and coarse to fine.}
The warm start fits an emission--absorption model~\cite{max1995optical} at $16^3$ for 10\,s, charged to every method's
clock.
The grid switches from $16^3$ to $32^3$ at 15\,s and to $64^3$ at 45\,s (PRB has its own schedules); at a switch,
$\theta$ is re-initialised from the upsampled density and Adam's moments are reset.

\paragraph{Iteration and clock.}
Each iteration draws 32768 rays uniformly from the pixel-centre rays of the training views that intersect the cube and
minimises their mean squared error of linear radiance with Adam~\cite{kingma2015adam} (learning rate 0.05,
$\epsilon=10^{-8}$).
A method's clock starts after a one-time warm-up of every model it uses, includes the warm start and the held-out
losses, and pauses during checkpoint evaluation; the GPU is synchronised before the clock is read.
Every run used one B200, one CPU core (four for PRB) and 6\,GB of host memory.

\FloatBarrier
\section{Gradient Decomposition}
\label{sec:supp:decomp}

\paragraph{Descent with a partial Jacobian.}
A step $\Delta\theta=-\eta\,\Jcam^\top\vr$ changes the loss by
$-\eta\,\tfrac{2}{P}\langle\mJ^\top\vr,\Jcam^\top\vr\rangle+O(\eta^2)$, which is $\cos(\Jcam^\top\vr,\mJ^\top\vr)$
times the change of a steepest-descent step of the same length: for a small step the frozen-source iteration descends
if and only if this cosine is positive.
Linearised at a consistent solution, its error evolves as
\begin{equation}
  \vdelta\leftarrow\big(\mathbf{1}-\eta\,\Jcam^\top\mJ\big)\,\vdelta ,
  \label{eq:linear}
\end{equation}
which decays along eigenvectors of $\Jcam^\top\mJ$ with eigenvalues of positive real part and grows along any with
negative real part, as for the unmatched projector--backprojector pairs of iterative
tomography~\cite{zeng2000unmatched,dong2019fixing}.
With consistent data every solution is a fixed point of every variant; with noise or model error the true gradient at
the frozen source's fixed point is $\tfrac{2}{P}(\Jdir+\Jms)^\top\vr$.

\begin{figure*}[t]
  \centering
  \includegraphics[width=\linewidth]{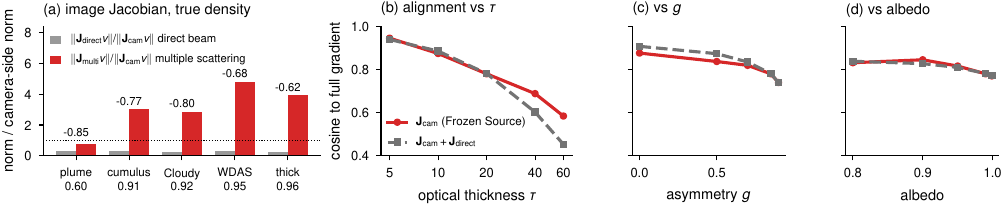}
  \caption{\textbf{The light-side derivative on development media} -- (a)~Norms of the direct-beam and
    multiple-scattering parts of the image Jacobian relative to the camera-side part (above the bars: the cosine between
    them; below the names: the multiply scattered fraction of the image). (b--d)~Cosine of the frozen-source gradient
    $\Jcam^\top\vr$ and of $(\Jcam+\Jdir)^\top\vr$ with the full gradient on the cumulus, against optical depth, $g$
    (albedo 0.99) and albedo ($g=0.85$). The multiple-scattering part dominates and grows with optical depth, and the
    frozen-source gradient loses its alignment accordingly.}
  \label{fig:supp:decomp}
\end{figure*}

\paragraph{Measurements.}
On five development media (model-consistent data rendered with the generic operator, 131072 rays of the 32 training
views) the Jacobian is measured along four perturbations $\vv$ of $\theta$ (a uniform scaling, two smooth random
fields and the layer within unit optical depth of the light), and the gradients at four perturbed densities (the truth
scaled by 0.8 and 1.25, and multiplied by two smooth log-normal fields); \cref{fig:supp:decomp} reports the means.
$\Jmsphi$, obtained by differentiating the operator, agrees with Monte Carlo central differences with cosine 0.94--0.99
in the 13 of 15 cases whose Monte Carlo noise is below 0.35.
$\|\Jms\vv\|/\|\Jcam\vv\|$ grows with optical depth (1.5 at $\tau=5$, 3.9 at $\tau=60$ with albedo 0.999), albedo and
$g$; $\Jdir$ and $\Jms$ do not cancel ($\|(\Jdir+\Jms)\vv\|$ is 5--21 times $\|\Jdir\vv\|$ on the clouds).
Adding $\Jdir$ without $\Jms$ does not help in clouds (the alignment is tied at $\tau\le20$ and lower by 0.09--0.13 at
$\tau\ge40$), consistent with the ablation of \cref{tab:supp:ablation}; on a thin 2D cloud slice a small gain from it was
reported~\cite{doicu2022surrogate}, and at $\tau=5$ our alignment does not separate the two either (0.95 and 0.94).

\paragraph{Source iteration and diffusion.}
Frozen-source differentiation lags the scattering source, just as source iteration does in transport solvers~\cite{adams2002fast}.
Diffusion synthetic acceleration approximates the missing response to accelerate transport~\cite{alcouffe1977diffusion,larsen1982unconditionally}; here $\Jmsphi$ supplies a learned response to the inverse problem instead.

\paragraph{P1 diffusion derivative.}
The classical control replaces the operator by the P1 approximation: for the current density a cell-centred
finite-volume P1 equation with Marshak boundaries and the single-scattered light as its source is solved by
Jacobi-preconditioned conjugate gradients; its degree-0 and degree-1 moments enter $S$ in place of the operator's, the
derivative flows through the solve by implicit differentiation, and the same anchor corrects the values every 30
iterations.
It runs 7.0--9.5 iterations per second (ours 17).
On eight development media it relights at 26.2--36.2\,dB after 2 and after 16\,min, against 40.4--49.0\,dB for ours
with the scope operator and 27.3--42.3\,dB for \FS after 2\,min; its selected iterate stops improving after
30--120\,s.
Diffusion is least accurate at the optically thin edges of clouds and for strongly forward-peaked scattering, which
these media have.

\paragraph{The learned field and derivative on held-out clouds.}
For each medium of the main evaluation (its $64^3$ truth, albedo, $g$ and training light; post hoc), a Gaussian bump
$\vv$ of $\log\sigt$ with a standard deviation of 3 voxels and peak 1 is placed at the density-weighted centroid, inside
the cloud (for medium 8 within two voxels of a gap, where 15\% of the bump falls outside the support and is masked).
The multiple-scattering source of light-side Monte Carlo at the truth (two independent halves of
\opfPhRefRange${}\times10^9$ photons; their difference puts the reference's noise below \opfRefNoiseMax) is compared
with each operator's prediction.
Its directional derivative along $\vv$ is estimated by central differences of $\log\sigt$ with $\epsilon=0.5$ and common
random numbers: photon-indexed random streams and one majorant for both densities, so that photon paths that never
meet the bump are identical and cancel; two independent pairs of \opfPhFdRange${}\times10^9$ photons per evaluation give
the estimate's own noise $r_n$.
The operators are evaluated in fp32 (the inversion runs them in bfloat16) and differentiated by central differences
with $\epsilon=0.02$ (within cosine \opfPrecMin{} of $\epsilon=0.05$); the trajectory operator's own difference at
$\epsilon=0.5$ lies within cosine \opfLinMin{} of this derivative, so its response is close to linear over this range,
and compared secant with secant the cosines below change by at most \opfSecantDiffMax.
Field errors and derivative cosines are $\sigs$-weighted over all coefficients (\cref{tab:supp:opfield}; cosines in
brackets corrected for $r_n$ as in \cref{tab:supp:jtraj}); the far field is the $\sigs$-weighted angular mean of the
derivative more than two bump standard deviations from its centre, with its share of the squared norm; image cosines
are over pixels.
The image changes of \cref{tab:supp:opfield} are rendered by our model from a camera at $25^\circ$ elevation, with the
light-side bracket of the unperturbed density held fixed (the frozen source, $\Jcam\vv$), with the operator's source and
the correction $\Delta$ of the unperturbed density ($(\Jcam+\Jdir+\Jmsphi)\vv$), and with the Monte Carlo sources of
both densities ($\mJ\vv$).
\Cref{fig:opderiv} shows the medium of \cref{fig:teaser} (\opfFigPhotonsRef{} reference photons in
both halves, \opfFigPhotonsFd{} per finite-difference evaluation); the main text quotes this run for that medium, the
table the uniform budget.

\begin{table*}[t]
  \centering\scriptsize\setlength{\tabcolsep}{3.5pt}
  \input{tables/opfield.tex}
  \caption{\textbf{The learned field and derivative on the main evaluation} -- Per medium ($\tau$: maximum column
    optical depth): value error of each operator at the truth, the trajectory operator's derivative along a local bump
    against the Monte Carlo central difference (cosine; in brackets corrected for its noise $r_n$), the far field of
    the derivative's angular mean (share of the squared norm, cosine), and in the image the share of the change's norm
    that the frozen-source derivative keeps and the cosine of ours with the Monte Carlo change. Uniform budget; post
    hoc.}
  \label{tab:supp:opfield}
\end{table*}

\FloatBarrier
\section{Operator Learning Analyses}
\label{sec:supp:oplearn}

\begin{table}[t]
  \centering\footnotesize
  \setlength{\tabcolsep}{2.5pt}
  \begin{tabular}{@{}lcccccc@{}}
    \toprule
    & gen. & scope & visited & \multicolumn{3}{c}{visited, by density error} \\
    \cmidrule(l){5-7}
    operator & val & val & states & $[0,1)$ & $[1,2)$ & $[2,3]$ \\
    \midrule
    generic    & 0.0228 & 0.0441 & 0.0725 & 0.045--0.057 & 0.079 & 0.119 \\
    generic+   & 0.0229 & 0.0420 & 0.0693 & 0.043--0.054 & 0.076 & 0.115 \\
    scope      & 0.0222 & 0.0337 & 0.0630 & 0.035--0.046 & 0.072 & 0.114 \\
    scope+     & 0.0214 & 0.0313 & 0.0597 & 0.032--0.044 & 0.068 & 0.109 \\
    trajectory & 0.0216 & 0.0315 & 0.0446 & 0.030--0.038 & 0.048 & 0.064 \\
    \bottomrule
  \end{tabular}
  \caption{\textbf{Field error by training data} -- Geometric mean of the relative $L_2$ error of the multiple-scattering
    source on the generic and scope validation sets and on the 95 held-back inversion states, also binned by the state's
    own density error ($[0,1)$: the bins $[0,0.5)$ and $[0.5,1)$; $n=12,28,36,19$). Development data.}
  \label{tab:supp:fielderr}
\end{table}

\begin{figure}[t]
  \centering
  \includegraphics[width=\linewidth]{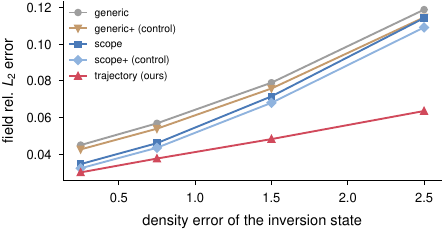}
  \caption{\textbf{Operator error on visited states} -- Field error of each operator on the 95 held-back inversion
    states against the state's own density error. The scope data lower the error near the truth; only the trajectory
    states lower it where the inversion drifts.}
  \label{fig:supp:oplearn}
\end{figure}

\paragraph{Data or training time.}
The scope data lower the field error on scope media by 24\% (generic+ by 5\%), and the trajectory states lower it on
the held-back inversion states by 29\% and on the most drifted ones by 44\% (scope+ by 5\% and 4\%;
\cref{tab:supp:fielderr,fig:supp:oplearn}).

\begin{figure}[t]
  \centering
  \includegraphics[width=\linewidth]{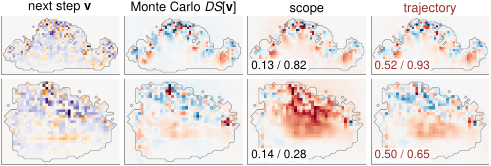}
  \caption{\textbf{Directional derivatives on two drifted states} -- The two held-back development states closest to
    both median next-step cosines of \cref{tab:supp:jtraj} (tied; density error 2.3): the inversion's next step $\vv$
    and the derivative of the multiple-scattering source along it (angular mean, vertical slice through the
    density-weighted centroid, one scale per row) from Monte Carlo ($4\times10^9$ photons per evaluation) and from
    the scope and trajectory operators; top and bottom row: the two states. Numbers: cosine with Monte Carlo over the
    whole field ($\sigs$-weighted, as in \cref{tab:supp:jtraj}) / of the displayed slices (unweighted).}
  \label{fig:supp:jtrajstates}
\end{figure}

\paragraph{Jacobian on visited states.}
Directional derivatives of the multiple-scattering source along a perturbation $\vv$ of $\log\sigt$ are compared by
central differences ($\epsilon=0.2$) for light-side Monte Carlo ($5\times10^8$ photons, common random numbers within a
pair; two independent pairs give the estimate's own noise $r_n$) and for each operator, on held-back states of the
trajectory data (per medium the stored states nearest density error 0.35, 1.0 and 2.2), along the inversion's next
step (to the next stored state), the drift (the state minus the truth in $\log\sigt$) and a smooth random field.
Errors and cosines are $\sigs$-weighted over all coefficients; errors are corrected for the Monte Carlo noise,
$e=\sqrt{\max(e_{\mathrm{raw}}^2-r_n^2,0)}$, and cosines by $\sqrt{1+r_n^2}$.
Along the next step and along the drift from drifted states, the derivatives of the generic, scope and scope+ operators
are nearly orthogonal or opposite to Monte Carlo's, and the trajectory operator's are not (\cref{tab:supp:jtraj}).
The weighting lets the density spikes of drifted states dominate: on the two states closest to both median next-step
cosines (a tie; \cref{fig:supp:jtrajstates}; recomputed post hoc with $4\times10^9$ photons per evaluation,
$r_n$ 0.04 and 0.06), voxels above $\sigt=500$, 0.3\% and 2\% of the support, carry 74\% and 85\% of the weighted energy of
the Monte Carlo derivative's angular mean.
Below $\sigt=500$ the trajectory operator is still closer (weighted angular-mean cosines 0.69 against 0.45 for scope,
0.62 against 0.28), though on the first state (top row) the two operators' slices look alike.

\begin{table*}[t]
  \centering\small
  \input{tables/jtraj.tex}
  \caption{\textbf{Directional derivatives on visited states} -- Median relative error / cosine of each operator's
    directional derivative against Monte Carlo on held-back inversion states, by direction and by the state's density
    error (noise-corrected). Development data.}
  \label{tab:supp:jtraj}
\end{table*}

\paragraph{In the loop (development).}
On three thick development media after 17500 iterations, the trajectory operator has current-iterate density errors
2.03 / 1.65 / 1.92, versus 8.13 / 9.51 / 3.80 with scope and 7.62 / 8.96 / 3.32 with the equal-steps scope+ control.
The corresponding anchor-estimated source errors are 43 / 21 / 33\% for trajectory and 208 / 219 / 83\% for scope.

\paragraph{Without selection (development).}
On eight development media with all 32 views and no selection (\cref{tab:supp:ns}), the trajectory operator relights
3.6\,dB better than the scope operator in the mean over 2--16\,min (8 of 8 media), 6.5\,dB at 16\,min, and diverges on
two media instead of six; it stays 1.9\,dB below the scope-operator pipeline with selection at 16\,min, which is why
selection remains part of the method.

\begin{table}[t]
  \centering\scriptsize\setlength{\tabcolsep}{2.3pt}
  \begin{tabular}{@{}rrrrrrr@{}}
    \toprule
    & selected & \multicolumn{2}{c}{final iterate} & & \multicolumn{2}{c}{traj., final $-$} \\
    \cmidrule(lr){3-4}\cmidrule(l){6-7}
    Budget & scope & traj. & scope & \FS & selected & scope \\
    \midrule
    30\,s   & 41.04 & 41.32 & 41.29 & 33.10 & $+0.28$ (6) & $+0.03$ (2)\\
    2\,min  & 44.12 & 44.52 & 44.07 & 34.08 & $+0.41$ (5) & $+0.45$ (6)\\
    8\,min  & 44.16 & 43.91 & 40.08 & 37.93 & $-0.24$ (5) & $+3.83$ (8)\\
    16\,min & 44.16 & 42.30 & 35.79 & 38.30 & $-1.86$ (2) & $+6.50$ (8)\\
    \bottomrule
  \end{tabular}
  \caption{\textbf{Trajectory training without selection} -- Development (eight media): mean relighting PSNR [dB] of the
    scope-operator pipeline with selection, of the trajectory and scope operators' final iterates on all 32 views, and of
    \FS (better prior); differences: mean over media (clipped at $\pm10$\,dB) and media ahead (of 8). Divergences
    (density error above 2): selected 0, trajectory 2, scope 6.}
  \label{tab:supp:ns}
\end{table}

\FloatBarrier
\section{Experimental Protocol}
\label{sec:supp:protocol}

\paragraph{Media.}
The main evaluation and the controlled set each hold six procedural cumulus clouds and six Cloudy
clouds~\cite{leonard2025ltdps} that no training
or development step used, each the first in a hash order of its candidates to pass an occupancy check, with maximum
column optical depth log-uniform in $[40,100]$, $1-\alpha$ log-uniform in $[10^{-3},10^{-2}]$, $g$ uniform in
$[0.7,0.9]$ and the development lights rotated about the vertical by a random angle.
The main evaluation's drawn values span $\tau$ 43.8--89.7, $\alpha$ 0.9901--0.9987 and $g$ 0.706--0.884 (controlled set: 42.8--99.8,
0.9909--0.999 and 0.703--0.888).
The main evaluation's media are $128^3$ volumes (procedural clouds generated at $128^3$, Cloudy clouds averaged to $128^3$) whose
$2^3$ averages are the $64^3$ truths; Mitsuba~3 renders its observations and references by volumetric path tracing
with Russian roulette disabled, and every inversion reads them from a cache that refuses any other source.
For the main evaluation the representation ceiling, our renderer's image of the $64^3$ truth against Mitsuba's references under the
new lights, is 44.7\,dB in the median (ours reaches 42.9\,dB after 16\,min), and the two renderers agree at $64^3$ to at
least 53.7\,dB.
The controlled set's observations and references are rendered by our path tracer from the $64^3$ truth, so that ours, \FS and the
\DO share the data's forward model up to noise (an inverse crime~\cite{kaipio2007inversecrimes} that favours them over
PRB).

\paragraph{Cameras, lights and metric.}
The 32 views are cameras on an orbit at distance 2.2 from the cube centre, evenly spaced in azimuth with elevations
cycling through $5^\circ$, $25^\circ$ and $45^\circ$, field of view $40^\circ$, $128^2$ pixels and 4096 samples per
pixel; four of them (every eighth) are held out for selection.
The training light is at an elevation of $60^\circ$; the three new lights are at $95^\circ$, $41^\circ$ and $59^\circ$
from it (elevations $22^\circ$, $22^\circ$, $53^\circ$), and two relighting cameras of $96^2$ pixels lie between
training views.
The metric is the noise-free relighting PSNR
\begin{equation}
  \mathrm{PSNR}=10\log_{10}\frac{\sum_{l,v}\max_p\big(R^{(1)}\big)^2}{\sum_{l,v}e_{lv}},
  \label{eq:supp:psnr}
\end{equation}
with $e_{lv}=\big\langle\big(A^{(1)}-R^{(1)}\big)\big(A^{(2)}-R^{(2)}\big)\big\rangle_p$
over the new lights $l$ and views $v$ and the mean over pixels $\langle\cdot\rangle_p$, where $A^{(k)}$ and $R^{(k)}$ are path-traced images (4096 samples per pixel) of
the reconstruction and of the truth with seed set $k$; the two seed sets are independent, so $e_{lv}$ has the
expectation of the squared error of the noise-free images.
Density errors are relative $L_2$ errors over the grid, and the projected optical depth error is that of the three
axis-aligned projections.

\paragraph{Baselines.}
\FS ran with two TV priors, $10^{-3}$ and $3\times10^{-3}$, each the best of its budgets on development media; the
tables report each, and the better per medium and budget (best of configuration).
The \DO uses its best development prior, TV $10^{-3}$; ours uses none.
PRB~\cite{vicini2021prb} runs in Mitsuba~3.9.1 with Dr.Jit~1.5.0 (integrator \texttt{prbvolpath}, unbounded paths);
the density comes from the shared parameterisation through \texttt{drjit.wrap}, and each iteration renders one random
training view with 64 samples per pixel for the primal and the adjoint pass.
Its settings were chosen per budget on four thick development media among the ten configurations of
\cref{tab:supp:prbpool2}, and the two longer-budget configurations (learning rates 0.01 and 0.005) were used in the original main evaluation.
We subsequently evaluated the previously development-selected short-budget configurations P30d (learning rate 0.16)
and P30b (learning rate 0.04) on all 12 main-evaluation media at 30\,s and 2\,min using the same main-evaluation
protocol, without Russian roulette. The reported 30\,s and 2\,min main-evaluation PRB values select the best of
these four fixed configurations per medium and budget; the 8- and 16-min values continue to use the original two.
This additional short-budget evaluation was performed after the original evaluation, without further hyperparameter tuning.
On the controlled set, the configurations tuned for 30\,s and 2\,min reached medians of 28.9 and 29.3\,dB at these budgets, against
24.2 and 28.2\,dB for the better of the two 30-min configurations.
On the 12-media main evaluation, P30d and P30b separately achieve 29.22 and 27.29\,dB at 30\,s and
29.43 and 29.76\,dB at 2\,min, respectively. Including the two original configurations, the per-medium
best-of-four PRB median is 29.30\,dB at 30\,s and 29.76\,dB at 2\,min. Ours leads this strengthened
PRB baseline on all 12 media, with median paired gains of 8.68\,dB [7.67, 9.75] and
12.18\,dB [10.72, 12.82] at the respective budgets (bootstrap 95\% intervals).
Without Russian roulette, which in media of albedo 0.99--0.998 reweights rare long paths by factors of
$10^2$--$10^4$, this baseline gains 1.3\,dB on the controlled set (post hoc), so the main evaluation runs it without; a TV prior changes PRB by
at most 0.2\,dB.
In PRB's favour are its tuning on development media and the best-of-configuration selection on the test media, and in
the main evaluation the observations come from its own renderer; in the controlled set they come from ours.

\begin{table}[t]
  \centering\scriptsize\setlength{\tabcolsep}{2.4pt}
  \input{tables/prb_pool2.tex}
  \caption{\textbf{PRB configurations} -- Mean relighting PSNR [dB] over the four thick development media of the choice;
    bold: the configuration chosen for each budget; $^\dagger$: the two configurations run to 30\,min; --: no checkpoint
    or a divergence. spp: samples per pixel; c2f: $32^3$ and $64^3$ stages from 15 and 45\,s (fast) or 60 and 240\,s
    (slow).}
  \label{tab:supp:prbpool2}
\end{table}

\begin{table}[t]
  \centering\small\setlength{\tabcolsep}{6pt}
  \begin{tabular}{@{}lrr@{}}
    \toprule
    Main-evaluation PRB configuration & 30\,s & 2\,min \\
    \midrule
    P30d (lr 0.16) & 29.22 & 29.43 \\
    P30b (lr 0.04) & 27.29 & 29.76 \\
    P120 (lr 0.01) & 24.36 & 28.36 \\
    P480 (lr 0.005) & 22.88 & 26.80 \\
    \midrule
    Best of four per medium & \textbf{29.30} & \textbf{29.76} \\
    \bottomrule
  \end{tabular}
  \caption{\textbf{PRB short-budget follow-up.} Main-evaluation median relighting PSNR [dB] across 12 media. P30d and P30b were added after the original evaluation using previously fixed development-set configurations. The best-of-four row selects per-medium maxima, not the maximum of the four medians.}
  \label{tab:supp:prbshort}
\end{table}

\paragraph{Statistics and failures.}
Comparisons are medians of per-medium paired differences with bootstrap 95\% intervals (the main evaluation's estimate of trajectory
against scope training: a mean with a $t$-interval).
A run whose density error exceeds 2 in an evaluated checkpoint, or whose relighting falls 1\,dB below its first
evaluation, counts as diverged and enters with its last evaluation; a job that ends early without diverging is rerun
once.

\paragraph{Pre-registration.}
For each held-out set the method, comparators, settings, claims, thresholds and failure rules were committed before
any of its media was generated, after independent read-only audits; the media were generated once, and the report ran
once on rows from the frozen code and checkpoints.
The main evaluation's thresholds are half the median effect on eight development media free of the inverse crime, rounded down to
a multiple of 0.5\,dB.
The controlled set tested ours with the scope operator, the method at the time, and the main evaluation the complete
recipe; every claim with a
threshold passed in both (\cref{tab:supp:p3claims}).

\begin{table*}[t]
  \centering\small\setlength{\tabcolsep}{4pt}
  \begin{tabular}{@{}l>{\raggedright\arraybackslash}p{0.47\linewidth}>{\raggedright\arraybackslash}p{0.15\linewidth}>{\raggedright\arraybackslash}p{0.22\linewidth}@{}}
    \toprule
    Claim & criterion on the 12 media of the main evaluation & development basis & main evaluation \\
    \midrule
    P1  & Ours $-$ \FS (better prior) after 2\,min: median $\ge$4\,dB and $\ge10$ ahead; after 16\,min: median
          $\ge$2.5\,dB & $+9.0$, 8; $+5.1$, 8 & $+6.8$, 12; $+4.0$, 12; pass \\
    P2  & Ours $-$ \DO after 2\,min: median $\ge$3.5\,dB & $+7.7$, 8 & $+10.2$, 12; pass \\
    P3  & Ours after 2\,min $\ge$ PRB (better configuration) after 30\,min on $\ge9$ & -- & 12 ($+9.0$); pass \\
    P4  & Ours $-$ Ours with the scope operator, mean over 2--16\,min (estimate) & $+1.3$ (4 Cloudy) &
          $+0.70$ [0.26, 1.14], 11 \\
    P5  & trajectory $-$ scope, final iterates after 8\,min: median $\ge$0.5\,dB and $\ge10$ ahead & $+1.6$, 8 &
          $+4.5$, 11; pass \\
    P5b & trajectory $-$ scope+, the same & $+1.8$, 8 & $+5.0$, 11; pass \\
    P6  & current density error after 16\,min against scope / scope+: lower on $\ge10$, median ratio $\le0.67$ &
          controlled set: 0.56, 11; 0.63, 12 & 0.28, 12; 0.41, 12; pass \\
    \bottomrule
  \end{tabular}
  \caption{\textbf{Pre-registered claims of the main evaluation} -- Median in dB and media ahead on the development basis
    (of 8) and on the main evaluation (of 12; P4: mean and 95\% $t$-interval, media positive). The controlled set's
    claims, for ours with the scope operator,
    all passed as well: $+6.6$\,dB over \FS (12 of 12), $+7.1$\,dB over the \DO (12), ahead of PRB after 30\,min on 12
    ($+12.0$\,dB), $+0.53$\,dB of the scope over the generic operator (9), and density-error ratios of the trajectory
    operator 0.56 (11) and 0.63 (12).}
  \label{tab:supp:p3claims}
\end{table*}

\FloatBarrier
\section{Additional Results}
\label{sec:supp:results}

\subsection{Main Evaluation}
\label{sec:supp:results:p3}

\begin{figure*}[t]
  \centering
  \includegraphics[width=\linewidth]{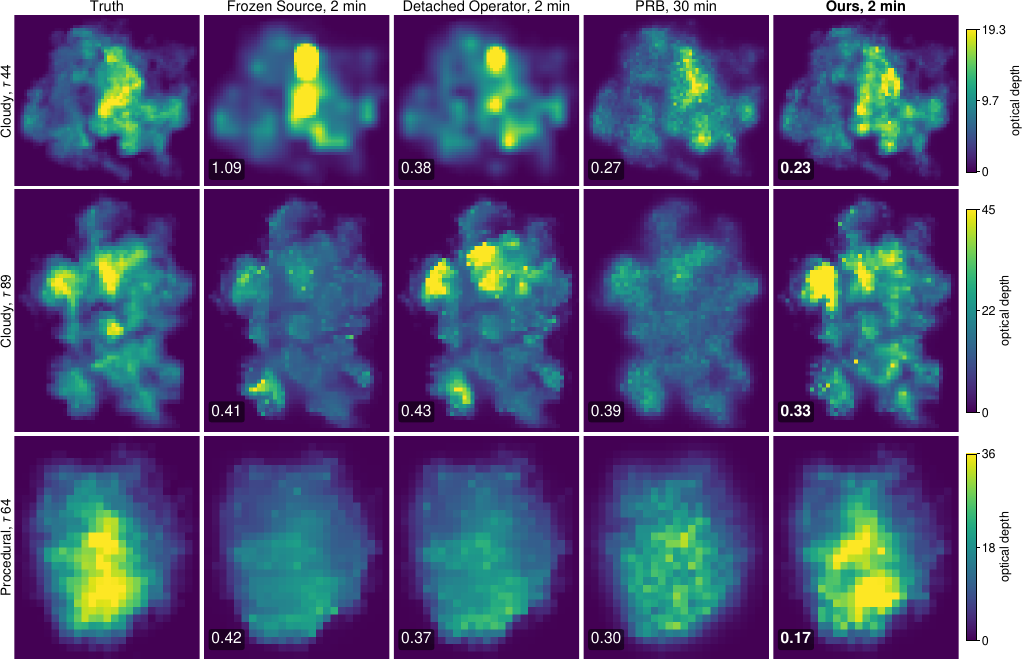}
  \caption{\textbf{Projected optical depth} -- Optical depth along the vertical (seen from above) of the truth and of the
    reconstructions for three media of the main evaluation, those at the 25th, 50th and 75th percentile of ours'
    advantage over the best baseline in projected optical depth (relative error over the three axis-aligned
    projections; ours lowest on 11 of 12). Colour bar per medium; number: relative error of the displayed projection,
    lowest in bold. The images constrain this quantity, and ours matches it best.}
  \label{fig:supp:od}
\end{figure*}

\Cref{tab:supp:p3sec} gives the paired comparisons at every budget, \cref{tab:supp:p3arms} iterations, density errors
and drift per method, \cref{tab:supp:p3pod,fig:supp:od} the projected optical depth, and
\cref{tab:supp:items,tab:supp:itemsfinal} the values per medium.
Ours leads \FS with either prior at every budget on all 12 media, and its lead over the better prior after 2\,min is
2.2--13.4\,dB per medium; the gain of the trajectory operator without selection is concentrated on Cloudy (7.5\,dB
after 8\,min, against 0.6\,dB on the procedural clouds, 5 of 6 positive).
PRB's better configuration reaches ours' 2-min value on no medium within 30\,min.
After 16\,min, ours' density error is the highest of the methods with selection (0.70 against 0.57--0.60), and its
projected optical depth error (0.24) is below that of \FS and the \DO (0.31--0.35) but above the scope operator's
(0.21).
Among the runs with selection, \FS with TV $10^{-3}$ diverged on 1 of 12 media by the density criterion and on 2
counting a 1-dB fall, the \DO on 2 counting a 1-dB fall, and every other method on none.

\begin{table}[t]
  \centering\scriptsize\setlength{\tabcolsep}{1.6pt}
  \input{tables/items_v2_final.tex}
  \caption{\textbf{Main evaluation, final iterates per medium} -- Relighting PSNR [dB] of ours (selected) and of the final
    iterates of the trajectory, scope and scope+ operators without selection. $^\dagger$: a run flagged as diverged.}
  \label{tab:supp:itemsfinal}
\end{table}

\begin{table*}[t]
  \centering\scriptsize\setlength{\tabcolsep}{3pt}
  \input{tables/pool3_secondary_cmp.tex}
  \caption{\textbf{Main evaluation, paired comparisons} -- Median paired difference in relighting PSNR [dB], bootstrap 95\%
    interval and media ahead at each budget. Final iterates: all 32 views, no held-out selection. PRB: the better of
    its original two longer-budget configurations at each budget. The subsequently added short-budget runs are summarized separately in \cref{tab:supp:prbshort}.}
  \label{tab:supp:p3sec}
\end{table*}

\begin{table*}[t]
  \centering\scriptsize\setlength{\tabcolsep}{3pt}
  \input{tables/pool3_secondary_arms.tex}
  \caption{\textbf{Main evaluation, per method} -- Medians over the 12 media: iterations reached, density error of the
    evaluated iterate, density error of the current iterate after 16\,min (in brackets: media above 2), divergences
    (density criterion) and the operator's error on the current density along the run (mean of the anchor error).}
  \label{tab:supp:p3arms}
\end{table*}

\begin{table}[t]
  \centering\scriptsize\setlength{\tabcolsep}{3pt}
  \input{tables/pool3_projected_od.tex}
  \caption{\textbf{Main evaluation, what is reconstructed} -- Median relative $L_2$ error of the density / of the projected
    optical depth of the evaluated iterate (PRB after 30\,min). Per medium, against \FS and the \DO after 2\,min and PRB
    after 30\,min (its configuration of better PSNR, lr 0.01 on all 12), ours after 2\,min has the lowest density error on 6
    of 12 (5 if lr 0.005 also counts) and the lowest projected-optical-depth error on 11 of 12.}
  \label{tab:supp:p3pod}
\end{table}

\begin{table*}[t]
  \centering\scriptsize\setlength{\tabcolsep}{3pt}
  \input{tables/items_v2_methods.tex}
  \caption{\textbf{Main evaluation, per medium} -- Relighting PSNR [dB] after 2\,min of ours with each operator, \FS with each
    prior and the better one, the \DO and PRB with each original longer-budget configuration and their better result, and of PRB's better
    configuration after 30\,min. $^\dagger$: a diverged run.}
  \label{tab:supp:items}
\end{table*}

\subsection{Controlled Set}
\label{sec:supp:results:p2}

\begin{table*}[t]
  \centering\scriptsize\setlength{\tabcolsep}{3pt}
  \input{tables/pool2_secondary_cmp.tex}
  \caption{\textbf{Controlled set, paired comparisons} -- Median paired difference in relighting PSNR [dB], bootstrap 95\%
    interval and media ahead. Ours: the scope operator. PRB: per medium the best configuration that reached the budget.}
  \label{tab:supp:p2sec}
\end{table*}

The controlled set ran ours with the scope operator and its variants (\cref{tab:supp:p2sec}), and
the ablations of \cref{sec:exp:ablations} (\cref{tab:supp:ablation}).
Ours leads each prior of \FS at every budget, and PRB over every configuration that reached the budget by
10.7--14.1\,dB; neither PRB configuration reaches ours' 2-min value within 30\,min on any medium.
With selection, the trajectory operator relights 0.44\,dB better than the scope operator in the mean over 2--16\,min,
the scope operator 0.53\,dB better than the generic one, and generic+ 0.13\,dB better than generic.
After the report, the projected optical depth error after 2\,min was 0.18 for both the trajectory and the scope
operator, against 0.40--0.48 for \FS, 0.43 for the \DO and 0.34--0.40 for PRB after 30\,min (post hoc).

\begin{table}[t]
  \centering\footnotesize\setlength{\tabcolsep}{3.2pt}
  \input{tables/ablation_v2.tex}
  \caption{\textbf{Ablations on the controlled set} -- Median relighting PSNR [dB] of our pipeline with the scope operator
    (the controlled set's method) when one component is removed (top), and of the baselines, which also differ in their
    prior (bottom; \FS: the better of two priors per medium). Median paired differences: \cref{tab:supp:p2sec}.}
  \label{tab:supp:ablation}
\end{table}

\subsection{Where the Learned Derivative Pays}
\label{sec:supp:regime}

The gain over the frozen source grows with optical depth.
On media of moderate optical depth, the scope-operator method and \FS are level (median $-0.01$\,dB after 2\,min on
five development media with $\tau$ 7.6--24, $-0.17$\,dB after 8\,min on six).
Inside the thick range of the controlled set the gain is largest on the eight media with $\tau\ge50$ (post hoc: median 7.8 /
6.2\,dB after 2 / 16\,min, against 1.5 / 1.0\,dB on the four with $\tau<50$).

\subsection{Limitations}
\label{sec:supp:limits}
The gains are limited to media in which multiple scattering dominates (in our tests maximum optical depth 40--100 and
albedo of at least 0.99); at moderate optical depth the frozen source is as accurate and cheaper
(\cref{sec:supp:regime}).
Training on the inversion's states slows the drift of the iterate but does not stop it, so the method selects its output
on held-out views (\cref{tab:supp:ns}).
We assume known albedo, phase function, cameras and a directional light, reconstruct at $64^3$ and test on synthetic
data.

\subsection{Compute}
Development (operator training, development inversions and analyses, baseline tuning) used about \gpuhDev{} B200
GPU-hours and the regime, controlled and main held-out sets \gpuhPoolOne{}, \gpuhPoolTwo{} and \gpuhPoolThree{}
(\gpuhTotal{} in total).

\paragraph{Online cost of the derivative.}
\Cref{tab:supp:dtime} times the derivative of \cref{fig:opderiv} (post hoc) with a synchronized wall clock on two B200s
of one node, each call on one GPU, from the density and $\vv$ in GPU memory to the derivative field (all coefficients);
loading the operator, warm-up calls and its offline training (once for all media: \dtTrain{} of network training for the
three stages on one B200, besides generating the training data) are excluded.
Ours is the central difference above, two fp32 forward passes with the optical depths and features (median of
\dtOursCalls{} calls, \dtOursRange); one forward pass takes \dtFwd{} in fp32 and \dtFwdBf{} in the inversion's bfloat16
(in a separate run, the inputs alone \dtInputs{} and the network alone \dtNetFp{} and \dtNetBf).
Monte Carlo is the central difference with common random numbers at \dtMinPhotons{} to \dtRefPhotons{} photons per
evaluation, with seeds independent of the reference's; the reference averages two central differences at
\dtRefPhotons{} (\dtRef{} in all), and \cref{fig:opderiv} shows the first of them.
Errors and cosines are taken against that reference and include its own noise (\dtRefNoise); the comparison is between
derivatives of the multiple-scattering field, not with PRB's gradient estimator.
Monte Carlo reaches our error (\dtOursErrThree) at \dtRefPhotons{} photons per evaluation (\dtMatchErr{} for its two
seeds), after \dtMatch, \dtMatchRatioExact{} times our time (\cref{fig:opderiv} rounds down), and is still less accurate
at \dtLbPhotons{} (\dtLb, \dtLbRatio{} times).
Its error is noise that falls with the photon count, ours a bias of the operator (it overshoots the lobe of the
response towards the light).
On this cloud the tracer is among the slowest (\dtRate{} photons per second; \dtRateRange{} over the clouds of the main
evaluation), so the ratio is likely smaller on most others; the cloud was chosen for \cref{fig:teaser}, not for its cost.

\begin{table}[!htbp]
  \centering\scriptsize\setlength{\tabcolsep}{3pt}
  \input{tables/dtime.tex}
  \caption{\textbf{Cost of the derivative of \cref{fig:opderiv}} -- Synchronized wall-clock time on one \dtGPU{} and
    the error (relative, $\sigs$-weighted, all coefficients) and cosine against the reference (two central differences
    with \dtRefPhotons{} photons per evaluation; its own noise \dtRefNoise). Monte Carlo: median time over seeds,
    range of the seeds' errors and cosines.}
  \label{tab:supp:dtime}
\end{table}

%% file: tables/op_ablation.tex
\begin{tabular}{@{}lccc@{}}
\toprule
Operator (30k steps) & field rel.\ $L_2$ & image PSNR & MS PSNR \\
\midrule
reference (U-Net) & 0.0411 & 52.52 & 48.00 \\
world-frame outputs & 0.0484 & 50.28 & 45.76 \\
source predicted directly (no analytic $g^\ell$) & 0.0419 & 52.46 & 47.94 \\
no forward-cascade features & 0.0432 & 52.08 & 47.56 \\
fixed offline set (4000 media) & 0.0369 & 53.50 & 48.98 \\
FNO backbone & 0.0537 & 49.95 & 45.43 \\
\bottomrule
\end{tabular}

%% file: tables/opfield.tex
\begin{tabular}{@{}rlrccccccccc@{}}
\toprule
& & & \multicolumn{3}{c}{value error} & \multicolumn{2}{c}{derivative (traj.)} & \multicolumn{2}{c}{far field} & \multicolumn{2}{c}{image change} \\
\cmidrule(lr){4-6}\cmidrule(lr){7-8}\cmidrule(lr){9-10}\cmidrule(l){11-12}
\# & family & $\tau$ & traj. & scope & generic & cosine & $r_n$ & share & cosine & \FS & ours, cos. \\
\midrule
0 & procedural & 89.1 & 0.025 & 0.027 & 0.043 & 0.95 (0.98) & 0.25 & 0.49 & 0.97 & 0.1\% & 0.91 \\
1 & procedural & 45.2 & 0.025 & 0.027 & 0.036 & 0.99 (1.00) & 0.13 & 0.32 & 0.99 & 1.2\% & 0.99 \\
2 & procedural & 51.1 & 0.031 & 0.033 & 0.042 & 0.98 (0.99) & 0.16 & 0.37 & 0.99 & 1.1\% & 0.99 \\
3 & procedural & 64.1 & 0.041 & 0.047 & 0.064 & 0.98 (0.99) & 0.12 & 0.25 & 0.99 & 1.5\% & 0.99 \\
4 & procedural & 50.3 & 0.032 & 0.037 & 0.047 & 0.97 (0.99) & 0.21 & 0.58 & 0.99 & 1.2\% & 0.98 \\
5 & procedural & 60.0 & 0.022 & 0.024 & 0.031 & 0.99 (1.00) & 0.14 & 0.28 & 0.99 & 1.2\% & 0.95 \\
6 & Cloudy & 49.4 & 0.027 & 0.032 & 0.042 & 0.99 (1.00) & 0.14 & 0.49 & 0.99 & 0.7\% & 0.99 \\
7 & Cloudy & 89.4 & 0.049 & 0.057 & 0.074 & 0.98 (0.99) & 0.17 & 0.40 & 0.99 & 1.6\% & 0.98 \\
8 & Cloudy & 89.7 & 0.039 & 0.043 & 0.057 & 0.99 (0.99) & 0.09 & 0.27 & 0.99 & 18.1\% & 0.99 \\
9 & Cloudy & 80.7 & 0.035 & 0.038 & 0.053 & 0.98 (1.00) & 0.15 & 0.28 & 0.99 & 1.6\% & 0.94 \\
10 & Cloudy & 43.8 & 0.027 & 0.029 & 0.041 & 0.98 (0.99) & 0.15 & 0.36 & 0.99 & 0.9\% & 0.99 \\
11 & Cloudy & 71.1 & 0.030 & 0.034 & 0.045 & 0.97 (0.99) & 0.21 & 0.31 & 0.98 & 14.8\% & 0.99 \\
\midrule
\multicolumn{3}{@{}l}{median} & 0.030 & 0.033 & 0.044 & 0.98 (0.99) & 0.15 & 0.34 & 0.99 & 1.2\% & 0.99 \\
\bottomrule
\end{tabular}

%% file: tables/jtraj.tex
\begin{tabular}{@{}llrcccc@{}}
\toprule
Direction & density error & $n$ & generic & scope & scope+ & trajectory \\
\midrule
next step & $[0,0.6)$ & 6 & $0.41$ / $0.91$ & $0.35$ / $0.93$ & $0.33$ / $0.94$ & $0.30$ / $0.95$ \\
 & $[0.6,1.5)$ & 10 & $1.01$ / $0.44$ & $1.03$ / $0.42$ & $1.03$ / $0.44$ & $0.84$ / $0.59$ \\
 & $\ge1.5$ & 8 & $1.41$ / $0.15$ & $1.49$ / $0.13$ & $1.43$ / $0.15$ & $0.87$ / $0.51$ \\
\multicolumn{7}{@{}l}{\quad trajectory below scope on 24/24 (median ratio 0.82), below scope+ on 24/24 (0.84)} \\
\midrule
drift & $[0.6,1.5)$ & 5 & $0.90$ / $0.37$ & $0.89$ / $0.38$ & $0.89$ / $0.39$ & $0.93$ / $0.37$ \\
 & $\ge1.5$ & 7 & $1.17$ / $-0.15$ & $1.20$ / $-0.16$ & $1.19$ / $-0.17$ & $0.92$ / $0.49$ \\
\multicolumn{7}{@{}l}{\quad trajectory below scope on 9/12 (median ratio 0.89), below scope+ on 9/12 (0.89)} \\
\midrule
smooth & $[0,0.6)$ & 6 & $0.19$ / $0.98$ & $0.14$ / $0.99$ & $0.13$ / $0.99$ & $0.12$ / $0.99$ \\
 & $[0.6,1.5)$ & 10 & $0.61$ / $0.79$ & $0.58$ / $0.81$ & $0.57$ / $0.82$ & $0.53$ / $0.85$ \\
 & $\ge1.5$ & 8 & $0.94$ / $0.48$ & $0.91$ / $0.52$ & $0.89$ / $0.54$ & $0.70$ / $0.74$ \\
\multicolumn{7}{@{}l}{\quad trajectory below scope on 22/24 (median ratio 0.87), below scope+ on 22/24 (0.91)} \\
\bottomrule
\end{tabular}

%% file: tables/prb_pool2.tex
\begin{tabular}{@{}lrrccrrrrr@{}}
\toprule
Config. & spp & lr & c2f & Adam $\epsilon$ & 30\,s & 120\,s & 480\,s & 960\,s & 1800\,s \\
\midrule
P16b & 16 & 0.04 & fast & $10^{-8}$ & 28.37 & 29.62 & 27.74 & -- & -- \\
P16 & 16 & 0.01 & slow & $10^{-8}$ & 26.26 & 29.29 & 30.64 & 30.24 & 29.40 \\
P30d & 64 & 0.16 & fast & $10^{-8}$ & \textbf{29.65} & 26.83 & -- & -- & -- \\
P30c & 64 & 0.08 & fast & $10^{-8}$ & 29.46 & 29.35 & 25.71 & -- & -- \\
P30b & 64 & 0.04 & fast & $10^{-8}$ & 28.72 & \textbf{30.84} & 29.18 & 27.28 & 25.32 \\
P30 & 64 & 0.02 & fast & $10^{-8}$ & 27.13 & 29.87 & 31.05 & 30.19 & 29.17 \\
P120 & 64 & 0.01 & slow & $10^{-8}$ & 25.60 & 29.83 & \textbf{32.22} & \textbf{32.92} & 32.70$^\dagger$ \\
P480 & 64 & 0.005 & slow & $10^{-16}$ & 23.78 & 27.88 & 30.64 & 31.72 & \textbf{32.80}$^\dagger$ \\
Pfb & 64 & 0.0025 & fast & $10^{-8}$ & 21.49 & 23.52 & 26.73 & 28.49 & 29.95 \\
P256 & 256 & 0.005 & slow & $10^{-16}$ & 22.11 & 25.31 & 29.02 & 30.75 & 32.48 \\
\bottomrule
\end{tabular}

%% file: tables/items_v2_final.tex
\begin{tabular}{@{}rlrrrrrrrrr@{}}
\toprule
& & & \multicolumn{4}{c}{8\,min} & \multicolumn{4}{c}{16\,min} \\
\cmidrule(lr){4-7}\cmidrule(l){8-11}
\# & family & $\tau$ & Ours & traj. & scope & scope+ & Ours & traj. & scope & scope+ \\
\midrule
0 & procedural & 89.1 & 41.40 & 41.65$^\dagger$ & 41.55$^\dagger$ & 41.03$^\dagger$ & 41.40 & 37.34$^\dagger$ & 30.79$^\dagger$ & 30.76$^\dagger$ \\
1 & procedural & 45.2 & 46.77 & 46.93 & 47.22 & 47.33 & 46.77 & 47.03 & 45.13 & 46.82 \\
2 & procedural & 51.1 & 45.07 & 45.10 & 44.32$^\dagger$ & 44.40$^\dagger$ & 45.30 & 45.22 & 29.77$^\dagger$ & 29.91$^\dagger$ \\
3 & procedural & 64.1 & 49.14 & 47.95 & 39.62$^\dagger$ & 41.18$^\dagger$ & 49.14 & 47.16 & 33.93$^\dagger$ & 34.58$^\dagger$ \\
4 & procedural & 50.3 & 42.55 & 43.22$^\dagger$ & 42.24$^\dagger$ & 42.66$^\dagger$ & 42.55 & 41.81$^\dagger$ & 32.50$^\dagger$ & 36.54$^\dagger$ \\
5 & procedural & 60.0 & 46.62 & 47.26 & 46.76$^\dagger$ & 46.48$^\dagger$ & 46.62 & 47.28 & 40.23$^\dagger$ & 39.42$^\dagger$ \\
6 & Cloudy & 49.4 & 43.26 & 43.97 & 40.33$^\dagger$ & 39.55$^\dagger$ & 43.26 & 43.32 & 35.53$^\dagger$ & 36.97$^\dagger$ \\
7 & Cloudy & 89.4 & 41.72 & 41.31$^\dagger$ & 35.87$^\dagger$ & 35.64$^\dagger$ & 41.72 & 37.36$^\dagger$ & 29.30$^\dagger$ & 29.25$^\dagger$ \\
8 & Cloudy & 89.7 & 41.69 & 42.72 & 35.20$^\dagger$ & 36.84$^\dagger$ & 41.69 & 42.67 & 30.09$^\dagger$ & 30.30$^\dagger$ \\
9 & Cloudy & 80.7 & 40.17 & 38.03$^\dagger$ & 26.56$^\dagger$ & 28.22$^\dagger$ & 40.17 & 30.71$^\dagger$ & 25.01$^\dagger$ & 25.23$^\dagger$ \\
10 & Cloudy & 43.8 & 43.43 & 44.24 & 34.78$^\dagger$ & 34.13$^\dagger$ & 43.43 & 43.92 & 27.97$^\dagger$ & 29.44$^\dagger$ \\
11 & Cloudy & 71.1 & 41.66 & 41.52 & 34.02$^\dagger$ & 33.96$^\dagger$ & 41.66 & 40.57 & 30.12$^\dagger$ & 30.94$^\dagger$ \\
\midrule
\multicolumn{3}{@{}l}{median} & 42.91 & 43.60 & 39.97 & 40.29 & 42.91 & 42.99 & 30.46 & 30.85 \\
\bottomrule
\end{tabular}

%% file: tables/pool3_secondary_cmp.tex
\begin{tabular}{@{}lcccc@{}}
\toprule
Comparison & 30\,s & 2\,min & 8\,min & 16\,min \\
\midrule
Ours $-$ \FS (TV $10^{-3}$) & +6.4 [5.0, 8.9] 12/12 & +10.7 [7.2, 13.9] 12/12 & +4.4 [3.0, 5.7] 12/12 & +4.2 [2.9, 5.2] 12/12 \\
Ours $-$ \FS (TV $3{\cdot}10^{-3}$) & +5.3 [4.9, 7.1] 12/12 & +6.9 [6.1, 10.7] 12/12 & +6.2 [3.6, 7.7] 12/12 & +6.3 [3.6, 7.8] 12/12 \\
Ours $-$ \FS (better prior) & +5.1 [4.7, 7.1] 12/12 & +6.8 [5.5, 10.7] 12/12 & +4.4 [2.4, 5.7] 12/12 & +4.0 [2.3, 5.1] 12/12 \\
Ours $-$ \DO & +5.2 [3.7, 7.1] 12/12 & +10.2 [5.3, 12.3] 12/12 & +5.9 [4.0, 8.8] 12/12 & +5.1 [3.4, 6.1] 12/12 \\
Ours $-$ Ours w/ scope op. & +0.0 [$-$0.1, 0.3] 7/12 & +0.5 [$-$0.1, 0.7] 9/12 & +0.9 [0.4, 1.5] 11/12 & +1.0 [0.4, 1.5] 11/12 \\
Ours $-$ Ours w/ scope+ op. & +0.1 [0.0, 0.3] 9/12 & +0.3 [0.0, 0.5] 9/12 & +0.7 [0.2, 1.0] 9/12 & +0.7 [0.1, 1.0] 9/12 \\
Ours (2\,min) $-$ PRB (better) & +17.6 [17.1, 18.3] 12/12 & +13.5 [12.9, 14.1] 12/12 & +10.9 [10.1, 11.9] 12/12 & +9.9 [9.1, 10.8] 12/12 \\
\midrule
traj.\ $-$ scope (final iterates) & $-$0.1 [$-$0.1, 0.2] 4/12 & +0.5 [0.2, 1.2] 10/12 & +4.5 [0.6, 7.9] 11/12 & +8.7 [6.8, 12.9] 12/12 \\
traj.\ $-$ scope+ (final iterates) & +0.0 [$-$0.3, 0.1] 6/12 & +0.3 [$-$0.2, 1.0] 8/12 & +5.0 [0.7, 7.2] 11/12 & +8.0 [5.9, 12.5] 12/12 \\
scope+ $-$ scope (final iterates) & +0.0 [$-$0.1, 0.1] 6/12 & +0.1 [$-$0.1, 0.5] 8/12 & +0.0 [$-$0.4, 1.0] 6/12 & +0.4 [0.1, 1.5] 9/12 \\
traj.\ (final iterate) $-$ Ours & +0.1 [$-$0.1, 0.2] 8/12 & +0.2 [0.0, 0.4] 9/12 & +0.2 [$-$0.3, 0.7] 8/12 & $-$0.4 [$-$3.0, 0.4] 5/12 \\
scope (final iterate) $-$ Ours w/ scope op. & +0.2 [0.0, 0.5] 9/12 & +0.0 [$-$0.2, 0.3] 7/12 & $-$3.3 [$-$6.9, 0.2] 3/12 & $-$10.6 [$-$14.2, $-$7.7] 0/12 \\
\bottomrule
\end{tabular}

%% file: tables/pool3_secondary_arms.tex
\begin{tabular}{@{}lccccccc@{}}
\toprule
Arm & \multicolumn{2}{c}{iterations} & \multicolumn{2}{c}{evaluated density error} & current, 16\,min & div. & anchor error \\
 & 2\,min & 16\,min & 2\,min & 16\,min & median (${>}2$) & & \\
\midrule
Ours (trajectory op.) & 1960 & 16180 & 0.55 & 0.70 & 1.49 (4) & 0 & 20\% \\
Ours w/ scope op. & 1992 & 15754 & 0.58 & 0.58 & 7.66 (11) & 0 & 67\% \\
Ours w/ scope+ op. & 1989 & 16198 & 0.53 & 0.60 & 5.54 (10) & 0 & 54\% \\
trajectory, final iterate & 2030 & 16897 & 0.56 & 1.32 & 1.32 (4) & 4 & 21\% \\
scope, final iterate & 1972 & 15924 & 0.60 & 7.35 & 7.35 (11) & 11 & 61\% \\
scope+, final iterate & 1952 & 16062 & 0.59 & 6.92 & 6.92 (11) & 11 & 59\% \\
\FS (TV $10^{-3}$) & 3355 & 35590 & 1.29 & 0.57 & 0.58 (0) & 1 & -- \\
\FS (TV $3{\cdot}10^{-3}$) & 3574 & 38020 & 0.65 & 0.57 & 0.56 (0) & 0 & -- \\
\DO & 2514 & 23283 & 0.98 & 0.58 & 0.59 (0) & 0 & 17\% \\
PRB, lr 0.005 & 460 & 2246 & 0.66 & 0.56 & 0.56 (0) & 0 & -- \\
PRB, lr 0.01 & 296 & 1562 & 0.59 & 0.56 & 0.56 (0) & 0 & -- \\
\bottomrule
\end{tabular}

%% file: tables/pool3_projected_od.tex
\begin{tabular}{@{}lcccc@{}}
\toprule
Arm & 2\,min & 8\,min & 16\,min & 30\,min \\
\midrule
Ours (trajectory op.) & 0.55 / 0.20 &  & 0.70 / 0.24 &  \\
Ours w/ scope op. & 0.58 / 0.21 &  & 0.58 / 0.21 &  \\
trajectory, final iterate &  & 0.81 / 0.31 & 1.32 / 0.49 &  \\
scope, final iterate &  & 2.05 / 0.65 & 7.35 / 3.61 &  \\
scope+, final iterate &  & 1.80 / 0.56 & 6.92 / 3.37 &  \\
\FS (TV $10^{-3}$) & 1.29 / 0.68 &  & 0.57 / 0.31 &  \\
\FS (TV $3{\cdot}10^{-3}$) & 0.65 / 0.42 &  & 0.57 / 0.33 &  \\
\DO & 0.98 / 0.53 &  & 0.58 / 0.35 &  \\
PRB, lr 0.005 &  &  &  & 0.55 / 0.36 \\
PRB, lr 0.01 &  &  &  & 0.58 / 0.31 \\
\bottomrule
\end{tabular}

%% file: tables/items_v2_methods.tex
\begin{tabular}{@{}rlrrrrrrrrrrrr@{}}
\toprule
& & & \multicolumn{3}{c}{Ours, 2\,min} & \multicolumn{3}{c}{\FS, 2\,min} & \DO & \multicolumn{3}{c}{PRB, 2\,min} & PRB \\
\cmidrule(lr){4-6}\cmidrule(lr){7-9}\cmidrule(lr){11-13}
\# & family & $\tau$ & traj. & scope & scope+ & $10^{-3}$ & $3{\cdot}10^{-3}$ & better & 2\,min & lr .005 & lr .01 & better & 30\,min \\
\midrule
0 & procedural & 89.1 & 40.42 & 39.56 & 40.22 & 27.22 & 27.40 & 27.40 & 27.14 & 24.40 & 25.98 & 25.98 & 28.41 \\
1 & procedural & 45.2 & 45.94 & 45.84 & 45.99 & 40.87 & 39.26 & 40.87 & 40.20 & 29.94 & 32.33 & 32.33 & 37.40 \\
2 & procedural & 51.1 & 43.53 & 43.41 & 43.51 & 41.34 & 38.72 & 41.34 & 40.85 & 29.47 & 31.24 & 31.24 & 36.73 \\
3 & procedural & 64.1 & 47.44 & 47.81 & 48.13 & 38.13 & 41.19 & 41.19 & 43.16 & 33.52 & 35.23 & 35.23 & 39.82 \\
4 & procedural & 50.3 & 43.38 & 43.71 & 43.18 & 39.03 & 39.40 & 39.40 & 38.61 & 25.69 & 27.74 & 27.74 & 32.93 \\
5 & procedural & 60.0 & 45.61 & 44.97 & 45.14 & 31.21$^\dagger$ & 37.22 & 37.22 & 33.69$^\dagger$ & 30.99 & 32.20 & 32.20 & 36.46 \\
6 & Cloudy & 49.4 & 42.59 & 42.09 & 42.24 & 27.65 & 29.20 & 29.20 & 29.52 & 27.23 & 28.78 & 28.78 & 33.82 \\
7 & Cloudy & 89.4 & 41.30 & 40.69 & 40.72 & 31.74 & 35.31 & 35.31 & 33.27 & 26.13 & 27.94 & 27.94 & 31.27 \\
8 & Cloudy & 89.7 & 41.05 & 40.04 & 40.30 & 30.60 & 33.95 & 33.95 & 29.57$^\dagger$ & 26.38 & 27.37 & 27.37 & 31.37 \\
9 & Cloudy & 80.7 & 40.17 & 39.60 & 39.77 & 25.82$^\dagger$ & 32.12 & 32.12 & 27.54 & 25.62 & 26.84 & 26.84 & 31.41 \\
10 & Cloudy & 43.8 & 42.47 & 42.74 & 42.84 & 29.10 & 29.41 & 29.41 & 30.77 & 28.86 & 29.96 & 29.96 & 35.04 \\
11 & Cloudy & 71.1 & 41.33 & 39.78 & 40.08 & 30.39 & 34.79 & 34.79 & 32.31 & 24.78 & 26.27 & 26.27 & 31.55 \\
\midrule
\multicolumn{3}{@{}l}{median} & 42.53 & 42.41 & 42.54 & 30.91 & 35.05 & 35.05 & 32.79 & 26.80 & 28.36 & 28.36 & 33.38 \\
\bottomrule
\end{tabular}

%% file: tables/pool2_secondary_cmp.tex
\begin{tabular}{@{}lcccc@{}}
\toprule
Comparison & 30\,s & 2\,min & 8\,min & 16\,min \\
\midrule
Ours $-$ \FS (TV $10^{-3}$) & +6.5 [2.3, 10.7] 11/12 & +8.4 [4.1, 15.2] 12/12 & +5.8 [1.8, 8.3] 12/12 & +5.7 [1.7, 7.0] 12/12 \\
Ours $-$ \FS (TV $3{\cdot}10^{-3}$) & +6.1 [3.5, 7.5] 12/12 & +7.6 [4.0, 9.2] 12/12 & +7.0 [3.3, 8.4] 12/12 & +6.8 [3.3, 8.3] 12/12 \\
Ours $-$ \FS (better prior) & +5.2 [2.3, 7.5] 11/12 & +6.6 [3.1, 9.2] 12/12 & +5.8 [1.5, 7.8] 12/12 & +5.7 [1.6, 7.0] 12/12 \\
Ours $-$ \DO & +5.5 [1.4, 9.7] 12/12 & +7.1 [2.3, 14.3] 12/12 & +6.1 [1.7, 8.4] 12/12 & +6.1 [1.5, 7.6] 12/12 \\
Ours $-$ Ours without $\Jdir$ & $-$0.1 [$-$0.3, 0.1] 4/12 & $-$0.1 [$-$0.2, 0.1] 5/12 & $-$0.3 [$-$0.4, 0.1] 5/12 & $-$0.3 [$-$0.4, 0.1] 5/12 \\
Ours $-$ Ours without anchor & $-$0.5 [$-$0.6, $-$0.3] 2/12 & +0.8 [0.4, 1.2] 10/12 & +0.9 [0.5, 1.3] 11/12 & +0.9 [0.5, 1.3] 11/12 \\
Ours $-$ PRB (best configuration) & +10.7 [10.4, 11.6] 12/12 & +14.1 [12.9, 15.0] 12/12 & +13.0 [12.3, 13.6] 12/12 & +12.3 [11.6, 13.0] 12/12 \\
\bottomrule
\end{tabular}

%% file: tables/ablation_v2.tex
\begin{tabular}{@{}lcccc@{}}
\toprule
Controlled set & 30\,s & 2\,min & 8\,min & 16\,min \\
\midrule
Ours (scope operator)                  & 39.3 & 43.5 & 43.7 & 43.7 \\
\quad without $\Jdir$                  & 39.4 & 43.5 & 44.1 & 44.1 \\
\quad without anchor ($\Delta=0$)      & 39.7 & 43.3 & 42.9 & 42.9 \\
\midrule
\DO (no $\Jdir{+}\Jmsphi$)             & 34.1 & 35.5 & 38.4 & 38.9 \\
\FS (no operator)                      & 34.7 & 36.9 & 38.0 & 39.1 \\
\bottomrule
\end{tabular}

%% file: tables/dtime.tex
\begin{tabular}{@{}lrcc@{}}
\toprule
Monte Carlo, photons / evaluation & time & error & cosine \\
\midrule
$10^{7}$ (3 seeds) & 1.9\,s & 3.60--3.61 & 0.264--0.271 \\
$3\times10^{7}$ (3 seeds) & 5.6\,s & 2.08--2.10 & 0.421--0.423 \\
$10^{8}$ (3 seeds) & 18\,s & 1.15 & 0.650--0.659 \\
$3\times10^{8}$ (3 seeds) & 55\,s & 0.662--0.667 & 0.828--0.832 \\
$10^{9}$ (3 seeds) & 3.0\,min & 0.371--0.381 & 0.934--0.937 \\
$3\times10^{9}$ (2 seeds) & 9.1\,min & 0.228--0.230 & 0.974--0.975 \\
$7.4\times10^{9}$ (2 seeds) & 23\,min & 0.162--0.163 & 0.987 \\
\quad reference: two at $7.4\times10^{9}$ & 45\,min & -- & -- \\
\midrule
Ours, two fp32 forward passes & 77\,ms & 0.163 & 0.987 \\
\quad one pass, fp32 / bfloat16 & 38\,ms / 18\,ms & & \\
\bottomrule
\end{tabular}